\documentclass[runningheads]{llncs}
\usepackage{graphicx}
\usepackage[inline]{enumitem}
\usepackage{float}
\usepackage{booktabs}
\usepackage{amsmath}
\usepackage{xcolor,colortbl}
\definecolor{babyblue}{rgb}{0.54, 0.81, 0.94}
\usepackage{multirow}
\usepackage{hyperref}
\begin{document}
%
\title{On the Trustworthiness of Tree Ensemble Explainability Methods}
%
%

\author{Angeline Yasodhara\thanks{These authors contributed equally to this work.} \and Azin Asgarian\textsuperscript{*} \and Diego Huang \and Parinaz Sobhani}

\authorrunning{A. Yasodhara et al.}
\institute{Georgian, 2 St Clair Ave West, Suite 1400
Toronto, Ontario, Canada M4V 1L5
\email{\{angeline,azin,diego,parinaz\}@georgian.io}\\
\url{https://georgian.io}}

\maketitle              
\vspace{-.5cm}
\begin{abstract}
The recent increase in the deployment of machine learning models in critical domains such as healthcare, criminal justice, and finance has highlighted the need for trustworthy methods that can explain these models to stakeholders. Feature importance methods (e.g. gain and SHAP) are among the most popular explainability methods used to address this need. For any explainability technique to be trustworthy and meaningful, it has to provide an explanation that is accurate and stable. Although the stability of local feature importance methods (explaining individual predictions) has been studied before, there is yet a knowledge gap about the stability of global features importance methods (explanations for the whole model). Additionally, there is no study that evaluates and compares the accuracy of global feature importance methods with respect to feature ordering.
In this paper, we evaluate the accuracy and stability of global feature importance methods through comprehensive experiments done on simulations as well as four real-world datasets. 
We focus on tree-based ensemble methods as they are used widely in industry and measure the accuracy and stability of explanations under two scenarios: 
\begin{enumerate*}
    \item when inputs are perturbed
    \item when models are perturbed.
\end{enumerate*} 
Our findings provide a comparison of these methods under a variety of settings and shed light on the limitations of global feature importance methods by indicating their lack of accuracy with and without noisy inputs, as well as their lack of stability with respect to:
\begin{enumerate*}
    \item increase in input dimension or noise in the data;
    \item perturbations in models initialized by different random seeds or hyperparameter settings.
\end{enumerate*} \\\\
This paper is a pre-published version of the original CD-MAKE 2021 publication: \url{https://doi.org/10.1007/978-3-030-84060-0\_19}.


\end{abstract}

\vspace{-.5cm}
\section{Introduction} \label{sec:intro}
\vspace{-.2cm}
Owing to the success and promising results achieved in supervised machine learning (ML) paradigm, there has been a growing interest in leveraging ML models in domains such as healthcare~\cite{asgarian2018hybrid,spann2020applying,yasodhara2021identifying}, criminal justice~\cite{rudin2019stop}, and finance~\cite{dixon2020machine}. As ML models become embedded into critical aspects of decision making, their successful adoption depends heavily on how well different stakeholders (e.g. user or developer of ML models) can understand and trust their predictions~\cite{asgarian2019limitations,christodoulakis2017barriers,doshi2017towards,lipton2018mythos,selbst2018intuitive}. As a result, there has been a recent surge in making ML models worthy of human trust~\cite{wiens2019no} and researchers have proposed a variety of methods to explain ML models to stakeholders~\cite{bhatt2020explainable}. 
Among these methods, feature importance methods in particular have received a lot of attention and gained tremendous popularity in industry~\cite{bhatt2020explainable}. The explanations obtained by these methods lie in two categories: \begin{enumerate*} \item local explanations \item global explanations \end{enumerate*}. Local explanations explain how a particular prediction is derived from the given input data. 
Global explanations, in contrast, provide a holistic view of what features are important across all predictions. 
Both explanation methods can be used for the purposes of model debugging, transparency, monitoring and auditing~\cite{bhatt2020explainable}. However, the trustworthiness and applicability of these explanations relies heavily on their accuracy and stability ~\cite{hancox2020robustness}.

Previously, Lundberg et al. \cite{NIPS2017_7062} assess the accuracy of feature importances by comparing them with human attributed importances. Ribeiro et al. \cite{ribeiro2016should} limits models to only use ten features from the input. Assuming the models would only pick the top ten important features, he then measures whether the selected features by the model are also captured by feature importances. Although both of these assessments capture whether important features are accurately identified, they do not measure the accuracy with respect to the relative ordering of features. We examine this with and without the presence of noisy inputs and use it to provide a comparison of different global feature importance methods.

In the explainability literature, various definitions are proposed for stability. Alvarez et al. \cite{alvarez2018robustness} define stability as being stable to local perturbations of the input, or in other words, similar inputs should not lead to significantly different explanations. Hancox-Li provides another definition for stability~\cite{hancox2020robustness}. He claims that stable explanations reflective of real patterns in the world are those that remain consistent over a set of equally well-performing models. Inspired by these definitions, we consider the following two scenarios to evaluate stability:  
\begin{enumerate*}
    \item local perturbations of the input
    \item perturbations of the models.
\end{enumerate*} 
We argue that stability with respect to these factors is essential to account for the inherent noisy nature of real-world data and to provide trustworthy explanations.

The stability of local explainability methods under the first scenario has been studied before. For example, Alvarez et al.~\cite{alvarez2018robustness} show LIME~\cite{ribeiro2016should} and (Kernel) SHAP~\cite{NIPS2017_7062} lack stability for complex black-box models through conducting the following experiments. They slightly perturb the input values and find that the surrogate models and original black box models produce stable output values whereas the explanations provided by LIME and SHAP change drastically in response to the perturbations. Despite these thorough investigations conducted on the stability of local explainability methods, there is yet little understanding about the stability of global explainability methods. With these methods getting embedded into critical aspects of daily life (healthcare, criminal justice, and finance)
, addressing this knowledge gap becomes crucial to avoid moral and ethical hazards~\cite{rudin2019stop}.

In this paper, we compare and evaluate the accuracy and stability of global feature explanation methods, gain and SHAP, through comprehensive experiments conducted on synthetic data and four real-world datasets. For this purpose, we use the following tree-based ensemble models as they are widely used in academia and industry: (1) random forest (2) gradient boosting machines~\cite{scikit-learn} and (3) XGBoost~\cite{Chen:2016:XST:2939672.2939785}. Our findings shed light on the limitations of the global explainability methods and show that they lack accuracy and become unstable when inputs or models are perturbed. 
For the rest of this paper, we first review the methodologies used in our experiments under Section \ref{sec:background}. We then describe our experimental setup in Section \ref{sec:exp_setup}. Finally, we present and discuss our findings in Sections \ref{sec:results} and \ref{sec:discussion} respectively.

\vspace{-.2cm}
\section{Background} \label{sec:background}
\vspace{-.2cm}
Tree ensemble methods are employed widely in research and industry due to their efficiency and effectiveness in modeling complex interactions in the data~\cite{biau2016random}. The two most common tree ensemble methods are gradient boosting \cite{friedman2002stochastic} and random forest \cite{breiman2001random}. In gradient boosting, trees are trained sequentially with upweighting the previously misclassified labels. In contrast, random forest trees are trained in parallel with different subsampling across all trees. We use random forest and gradient boosting machine implemented by sklearn~\cite{scikit-learn}, as well as XGBoost, a faster version of gradient boosting that uses second-order gradients~\cite{Chen:2016:XST:2939672.2939785}.

In this study, to compute global feature importances in tree ensemble methods we use gain~\cite{friedman2001elements} and SHAP~\cite{NIPS2017_7062}, an implementation of the Shapley algorithm. We focus on SHAP instead of LIME~\cite{ribeiro2016should} as LIME explanations can be fragile due to sampling variance \cite{bhatt2020explainable} and less resilient against adversarial attacks as shown by~\cite{slack2020fooling}. 
In the following sections, we briefly explain how gain~\cite{friedman2001elements} and SHAP~\cite{NIPS2017_7062} are computed.

\paragraph{Gain.}
\label{subsec:default_imp}
For both of the aforementioned tree ensemble methods, sklearn \cite{scikit-learn} and xgboost \cite{Chen:2016:XST:2939672.2939785} libraries provide the implementation to obtain the feature importances based on Hastie's description in the Elements of Statistical Learning \cite{friedman2001elements}. This is also referred to as \emph{gain}. This metric represents the improvements in accuracy or improvements in decreasing uncertainty (or variance) brought by a feature to its branches. At the end, to get a summary of the whole tree ensemble, this measure is averaged across all trees \cite{friedman2001elements,lewinson_2020,abu_2019}. 
In this paper, for the sake of simplicity and consistency we refer to this method as \emph{gain}.

\paragraph{SHAP.} \label{subsec:shap_imp}
SHapley Additive exPlanations (SHAP)~\cite{NIPS2017_7062} has gained a lot of attention in industry as a way to measure feature importance~\cite{bhatt2020explainable}. SHAP is an implementation of Shapley formula that summarizes the contribution of a feature to the overall prediction by approximating the Shapley value presented in the following:
\begin{equation*}
\phi_i = \sum_{S \subseteq F \backslash\{i\}} \frac{|S|!(|F| - |S| - 1)!}{|F|!}
[f_{S\cup\{i\}}(X_{S \cup \{i\}}) - f_S(X_S)]
\end{equation*}
where $\phi_i$ is the Shapley value for feature $i$, $S$ is a subset of all features $F$ that does not include feature $i$, $f_{S\cup\{i\}}$ is the model trained on features in $S$ and feature $i$, $f_S$ is the model trained on features in $S$, and $X$ is the input data. 

SHAP inherently calculates local importances, i.e. how each feature contributes to the prediction of a specific input. By averaging the absolute value of these local importances across the training set, one can obtain a global summary of how the feature as a whole contributes to the model. 
In this paper, we investigate the accuracy and stability of Tree SHAP~\cite{lundberg2020local2global} (a recent extension to Kernel SHAP with faster computation runtime for trees) under various settings. Unlike Kernel SHAP~\cite{NIPS2017_7062} which uses perturbation, Tree SHAP (with tree\_path\_dependent setting) leverages trees' cover statistics for fast approximation of Shapley values. 

\vspace{-.2cm}
\section{Experimental Setup} \label{sec:exp_setup}
\vspace{-.2cm}
In this section we describe the setup we use to evaluate the accuracy and stability of global feature importance methods.


\paragraph{Datasets.} To thoroughly evaluate the accuracy and stability of global feature importances, we conduct our experiments on synthetic data as well as four real-world datasets from various domains.

For synthetic data, we generate 300 training samples with varying number of features (5, 10, 25, 100, and 150 features). 
We randomly set the features to be either continuous or categorical (each with equal probability). For continuous features, we sample from a uniform distribution between $[0, 1)$. For categorical features, we first randomly sample values like continuous features and we then binarize them based on an independently-sampled threshold selected from $[0, 1)$. Lastly, to obtain the target values, we sum the multiplication of each feature by a randomized set of coefficients (sampled independently per feature between -10 to 10). Then, we categorize the summation values to 1 for values greater than the median and 0 otherwise. 

We use the following four real-world datasets in addition to the synthetic data for our stability assessments:
\begin{enumerate}
    \item Forest Fire: prediction of the amount of burned area resulted from forest fires in the northeast region of Portugal, by using meteorological data, such as coordinates, time, wind, rain, relative humidity, etc. \cite{cortez2007data}.
    \item Concrete: prediction of concrete compressive strength given material types, composition, and age \cite{yeh1998modeling}.
    \item Auto MPG: prediction of fuel consumption in miles per gallon (MPG) of cars in the city given its model, horsepower, etc. \cite{quinlan1993combining}
    \item Company Finance\footnote{This dataset is confidential and the details of it cannot be shared.}: prediction of whether companies would make a good investment based on their finances.
\end{enumerate}

All datasets except the Company Finance dataset (our proprietary dataset) come from the UCI ML data repository~\cite{Dua:2019} and are parsed with the py\_uci package~\cite{skafte_2019}. A summary of these datasets is shown in Table \ref{tab:dataset}. 
                        
\begin{table}
 \centering
  \caption{Description of datasets used in this study.}
  \label{tab:dataset}
  \scriptsize
  \begin{tabular}{ccccc}
    \toprule
    Dataset & Domain & Task Type & \# Samples & \# Features \\
    \midrule
    Synthetic Data & & Classification & 300 & 5-150 \\
    Forest Fire & Meteorology & Regression & 517 & 12 \\
    Concrete & Civil & Regression & 1030 & 8 \\
    Auto MPG & Automotive & Regression & 406 & 7 \\
    Company Finance & Finance & Classification & 2716 & 892 \\
  \bottomrule
\end{tabular}
\end{table}

\paragraph{Experimental Settings.}
In our experiments, we use random forest and gradient boosting machine implemented by sklearn package~\cite{scikit-learn}, as well as XGBoost, an implementation of gradient boosting that uses second-order gradients and has a faster runtime~\cite{Chen:2016:XST:2939672.2939785}. For each of these models, we run the following experiments:
\begin{enumerate}
    \item Input perturbation: where the input data are perturbed by adding different levels of noise. Noise is sampled randomly from a normal distribution with mean 0 and standard deviation of:
    \begin{enumerate*}
        \item half of the original feature's standard deviation for low noise
        \item the original feature's standard deviation for medium noise
        \item double of the original feature's standard deviation for high noise.
    \end{enumerate*}
    \item Model perturbation: where the model is perturbed by \begin{enumerate*}
        \item initializing with a different random seed without hyperparameter tuning, or
        \item optimizing hyperparameters~\cite{bergstra2013making} (e.g., number of trees, depth of trees, etc.) with a different random seed.
    \end{enumerate*} In these experiments, we ensure that the predictions of the two models (the original model and the perturbed model) have high correlations, such that of discrepancies in predictions affect the analysis minimally. 
\end{enumerate}

We iterate all experiments 50 times with a different random seed, except for the Company Finance dataset. For this dataset, we run the experiments 5 times due to long training time caused by the high number of features. In each iteration of input perturbation experiment, we train two models, one with the original setting (e.g., unperturbed input data) and another with the perturbed setting (e.g., noised input data). In model perturbation experiments, we also train two models in each iteration where we change the random seed of the second model to be different than the first model. For each trained model, we compute gain and SHAP feature importances as described in Section \ref{sec:background}. 


\paragraph{Accuracy Metrics.}
To evaluate the accuracy of global feature importances, we use simulated data so that the true coefficients (importances) are known. The features are ranked based on the magnitude of their corresponding coefficients used during data generation. 
We examine the accuracy under the following scenarios: \begin{enumerate*}
    \item when no noise is added to the input, and
    \item when different level of noise is added to the input.
\end{enumerate*}
We do not consider the model perturbation scenario for this analysis as we are mainly interested in measuring the accuracy of the model's feature importances to the true coefficients.

We evaluate the accuracy of the top features' ranking in the following way:
\begin{itemize}
    \item First, we rank features based on their coefficients' magnitude, largest magnitude being the most important. Since all features are uniformly sampled from $[0, 1)$, we assume the coefficients' magnitude represent the importances. 
    \item Second, we assess whether these top features are ranked correctly with gain and SHAP feature importances. 
    \item Finally, we count the number of times each top feature is ranked correctly by gain or SHAP feature importances across multiple iterations. If it is ranked incorrectly, there are 2 possible situations: \begin{enumerate*}
        \item The feature is still considered a top feature by gain or SHAP feature importances,
        \item The feature is not considered a top feature by gain or SHAP feature importances.
    \end{enumerate*}
    We present this count proportionally across the 3 groups (correct, incorrect\_but\_top, and incorrect) to compare the accuracy of these models on different levels.
\end{itemize} 
Furthermore, to get a sense of feature importances' accuracy across all features, we evaluate the Spearman correlation of gain and SHAP feature importances compared to the coefficients.

\paragraph{Stability Metrics.}
To evaluate the stability of global feature importances, we consider the following two scenarios:
\begin{enumerate*}
    \item when different levels of noise is added to the input.
    \item when models are perturbed by initializing with different random seeds and different hyperparameter settings.
\end{enumerate*}
We use Spearman correlation to compare feature importances calculated from the 2 models (one unperturbed and the other perturbed), because it is distribution-free unlike parametric tests (e.g., Pearson correlation)~\cite{zwillinger1999crc}. We also report both the Spearman and Pearson correlations between the predicted outputs of the two models trained in each iteration as a sanity check to ensure similar performance.



\vspace{-.3cm}
\section{Results} \label{sec:results}
\vspace{-.3cm}
Here, we present our findings from the experiments described in Section~\ref{sec:exp_setup}. We first discuss the accuracy of gain and SHAP feature importances in Section \ref{results:accuracy}. We then dive into the stability of each feature importance method when inputs are perturbed and when models are perturbed in Section \ref{results:stability}. Finally, we present a summary of our findings in Section \ref{results:summary}.

\vspace{-.3cm}
\subsection{Accuracy of Gain and SHAP Feature Importances}
\label{results:accuracy}
Table \ref{tab:correctness} demonstrate the accuracy of gain and SHAP for the top 3 features in synthetic data with a total of 5 features trained with XGBoost. The difference between SHAP and gain proportions are  highlighted beneath them. Orange indicates SHAP having a higher proportion and vice versa for blue. Models included in this experiment are highly predictive, with an average area under receiver operating curve (AUROC) of 92.6\% with standard deviation of 0.8\%.

Surprisingly, we find that the number of features ranked correctly is quite low for both methods even when there is no noise added to the input. For example, the rank \#1  feature is correctly ranked approximately 40\% of the time by both methods. Despite both SHAP and gain calculating feature importances from the same model, SHAP shows a slightly higher accuracy in ranking top features especially when noise is added into the input.


\begin{table*}[t]
\centering
  \caption{Proportions of correct, incorrect\_but\_top, and incorrect ranking of the top 3 features on synthetic data (total features: 5) using XGBoost model across all experiment iterations. Proportions in each column add up to 1. Highlighted values indicate the difference between SHAP and gain proportions: orange when SHAP having higher proportion and blue otherwise.}
  \label{tab:correctness}
  \scriptsize
\begin{tabular}{l|llllll|llllll}
    \toprule
  Experiment setting:                    & \multicolumn{6}{c|}{No noise added to   input}                                     & \multicolumn{6}{c}{Low noise added to   input}                                    \\
Original feature rank: & \multicolumn{2}{c}{1}     & \multicolumn{2}{c}{2}     & \multicolumn{2}{c|}{3}     & \multicolumn{2}{c}{1}     & \multicolumn{2}{c}{2}     & \multicolumn{2}{c}{3}     \\
Feature importance method:                      & gain         & shap       & gain        & shap        & gain        & shap        & gain         & shap       & gain         & shap       & gain          & shap      \\
                      \midrule
\multirow{2}{*}{correct}               & 0.44         & 0.5        & 0.46        & 0.44        & 0.32        & 0.3         & 0.44         & 0.46       & 0.4          & 0.52       & 0.3           & 0.44      \\
                      & \multicolumn{2}{c}{\cellcolor{orange}0.06}  & \multicolumn{2}{c}{\cellcolor{babyblue}-0.02} & \multicolumn{2}{c|}{\cellcolor{babyblue}-0.02} & \multicolumn{2}{c}{\cellcolor{orange}0.02}  & \multicolumn{2}{c}{\cellcolor{orange}0.12}  & \multicolumn{2}{c}{\cellcolor{orange}0.14}  \\
\multirow{2}{*}{incorrect\_but\_top}   & 0.26         & 0.28       & 0.22        & 0.26        & 0.32        & 0.28        & 0.24         & 0.16       & 0.26         & 0.2        & 0.46          & 0.38      \\
                      & \multicolumn{2}{c}{\cellcolor{orange}0.02}  & \multicolumn{2}{c}{\cellcolor{orange}0.04}  & \multicolumn{2}{c|}{\cellcolor{babyblue}-0.04} & \multicolumn{2}{c}{\cellcolor{babyblue}-0.08} & \multicolumn{2}{c}{\cellcolor{babyblue}-0.06} & \multicolumn{2}{c}{\cellcolor{babyblue}-0.08}  \\
\multirow{2}{*}{incorrect}             & 0.3          & 0.22       & 0.32        & 0.3         & 0.36        & 0.42        & 0.32         & 0.38       & 0.34         & 0.28       & 0.24          & 0.18      \\
                      & \multicolumn{2}{c}{\cellcolor{babyblue}-0.08} & \multicolumn{2}{c}{\cellcolor{babyblue}-0.02} & \multicolumn{2}{c|}{\cellcolor{orange}0.06}  & \multicolumn{2}{c}{\cellcolor{orange}0.06}  & \multicolumn{2}{c}{\cellcolor{babyblue}-0.06} & \multicolumn{2}{c}{\cellcolor{babyblue}-0.06} \\
                      \bottomrule
\end{tabular}
\end{table*}
 
\begin{figure*}[h]
  \centering
  \vspace{0.2cm}
  Model: \hspace{1cm} XGBoost \hspace{1.5cm} Gradient Boosting Machine \hspace{0.7cm} Random Forest \\ 
  \includegraphics[width=0.09\linewidth]{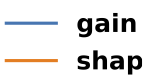}
  \includegraphics[width=0.29\linewidth]{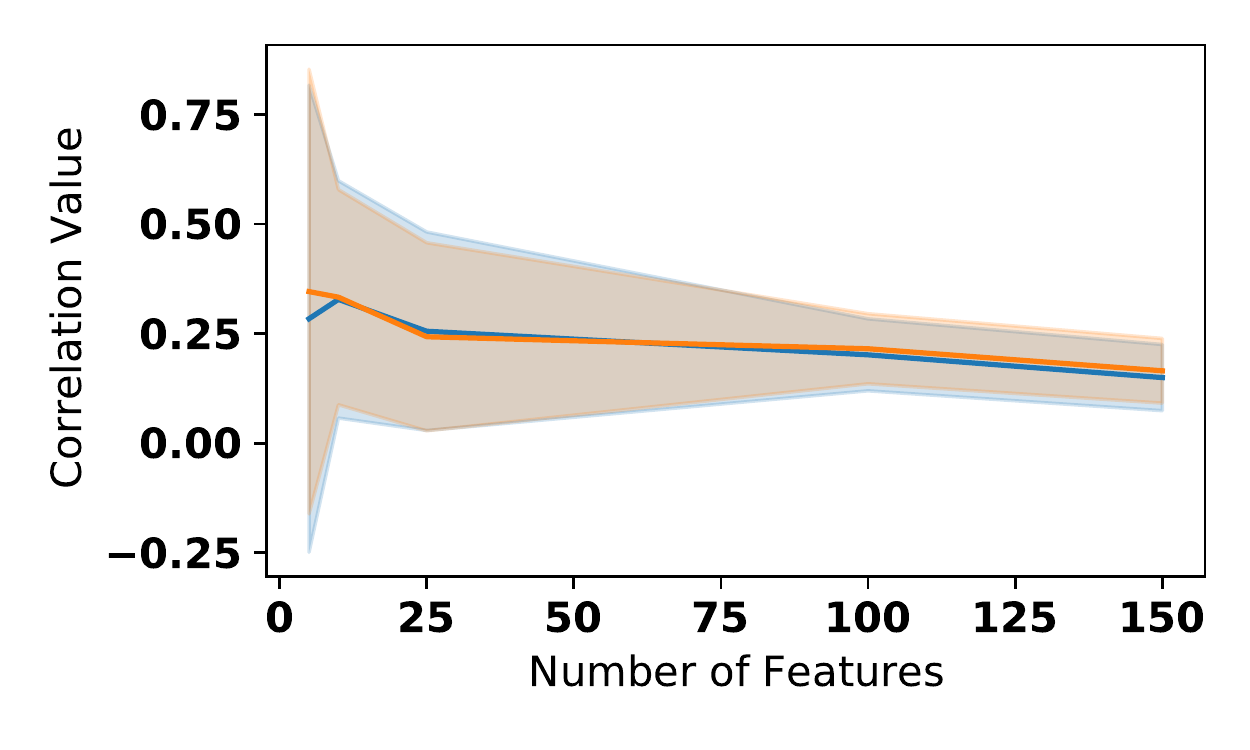}
  \includegraphics[width=0.29\linewidth]{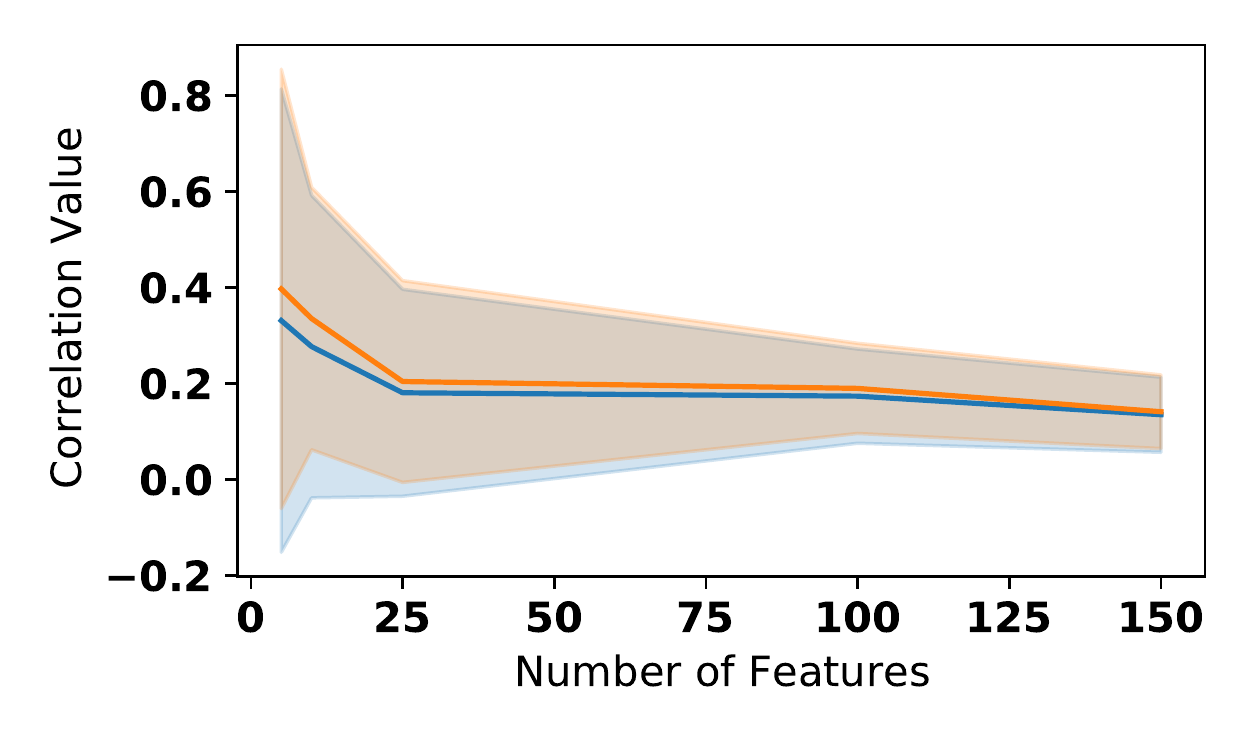}
  \includegraphics[width=0.29\linewidth]{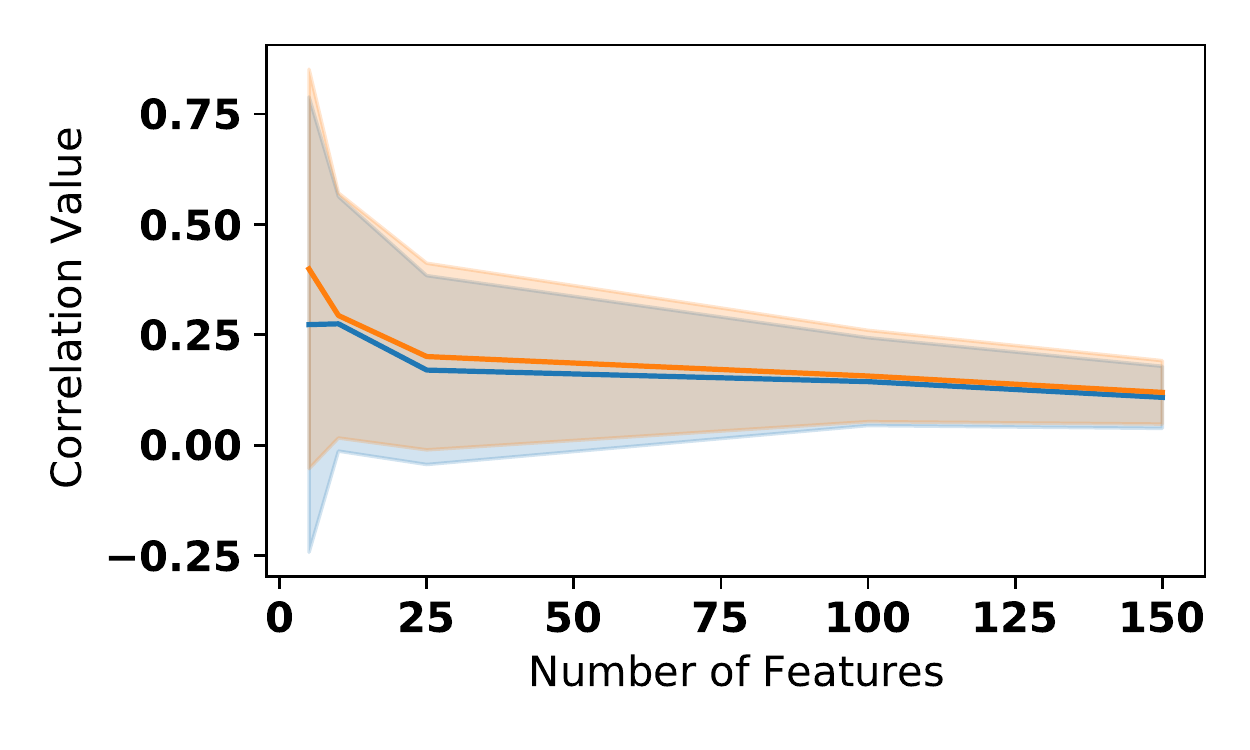}

  \caption{Spearman correlation of gain and SHAP feature importances (Blue: gain, Orange: SHAP) with the true coefficients with no noise added in simulation. Correlation is quite low across all settings.}
  \label{fig:correctness_corr}
\end{figure*}

To explicitly look at whether the feature importances provides an accurate ranking of all features, we further examine the Spearman correlation between the feature importances and the true coefficients. Figure \ref{fig:correctness_corr} shows the correlations in a noise-free scenario with increasing number of features. As demonstrated in this Figure, we find that gain and SHAP feature importances do not correlate well with the true coefficients (correlations range from 30-40\% and drops to around 20\% as the number of features increases). We observe a similar pattern across all other experimental settings (low-noised, medium-noised, or high-noised input).


\vspace{-.3cm}
\subsection{Stability of Gain and SHAP Feature Importances}
\label{results:stability} 
\begin{figure}[h]
  \centering
   \includegraphics[width=0.2\linewidth]{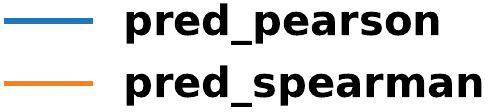} \\
  Perturbation: Random seeds \\
     XGBoost  \hspace{1.7cm} Gradient Boosting Machine \hspace{1cm} Random Forest \\ 
  \includegraphics[width=0.31\linewidth]{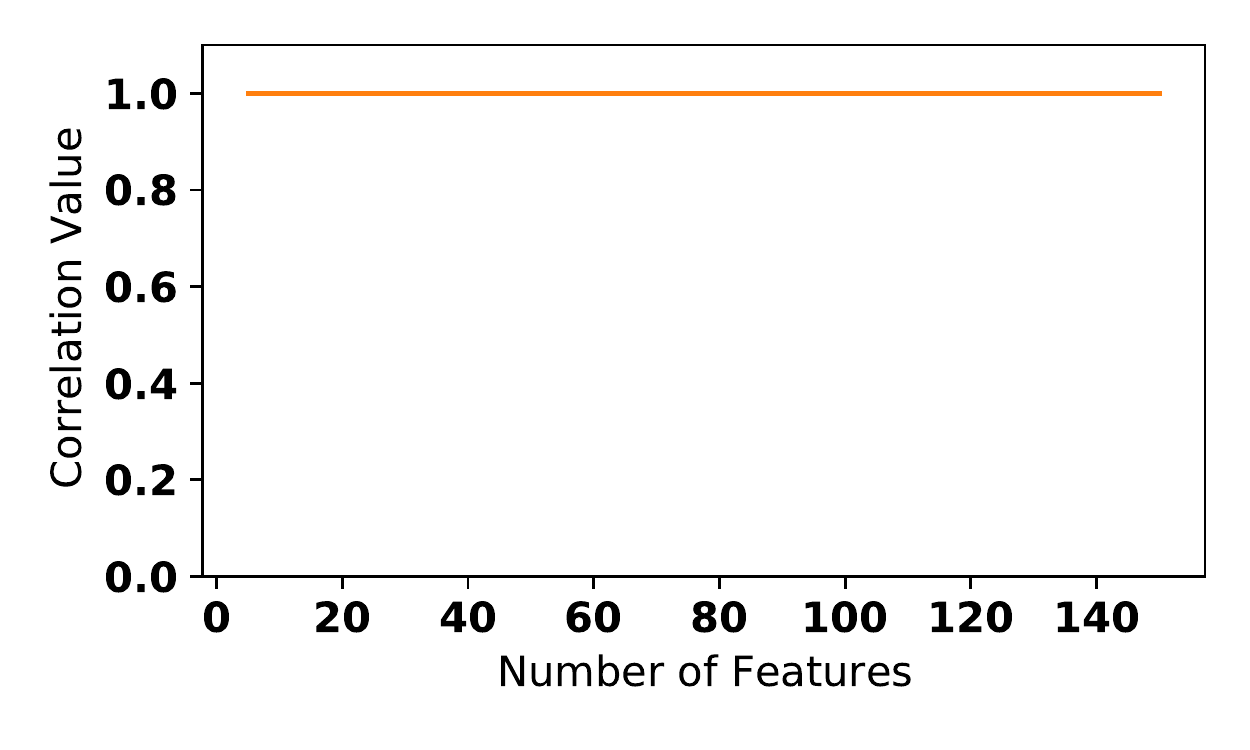}
   \includegraphics[width=0.31\linewidth]{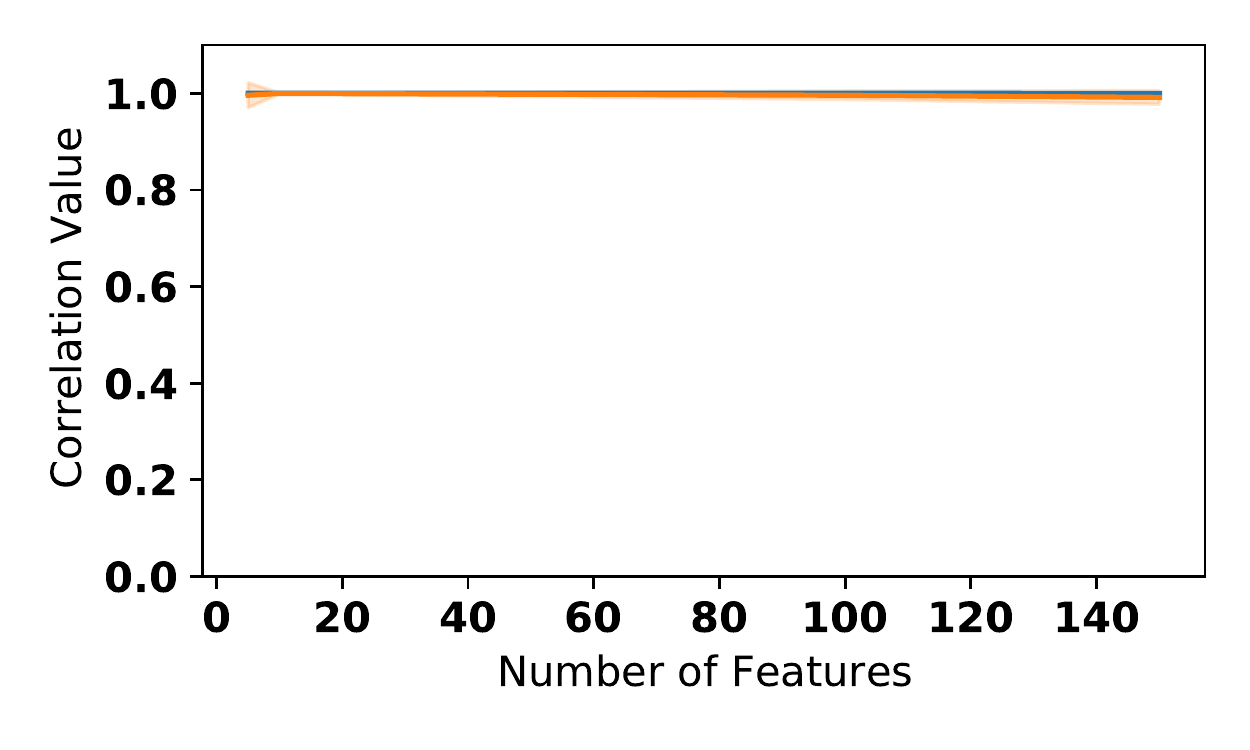}
    \includegraphics[width=0.31\linewidth]{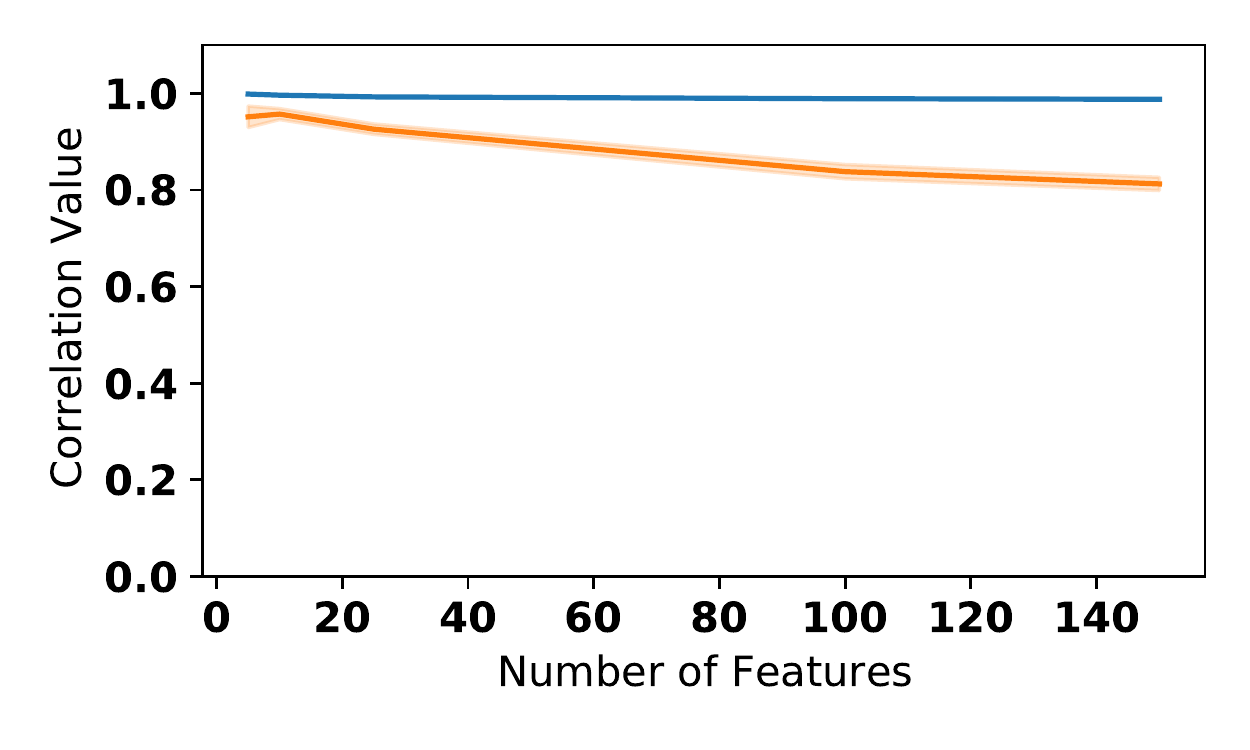}
      Perturbation: Hyperparameter settings \\
     XGBoost  \hspace{1.7cm} Gradient Boosting Machine \hspace{1cm} Random Forest \\ 
  \includegraphics[width=0.31\linewidth]{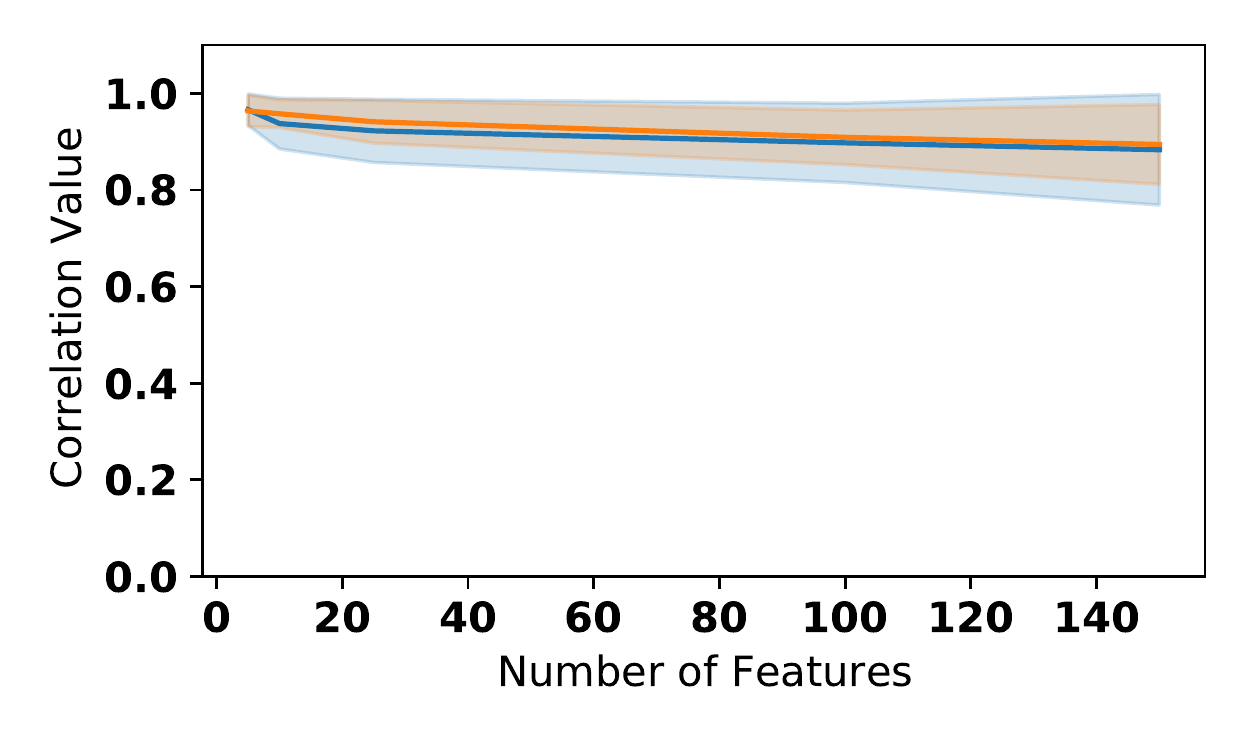}
   \includegraphics[width=0.31\linewidth]{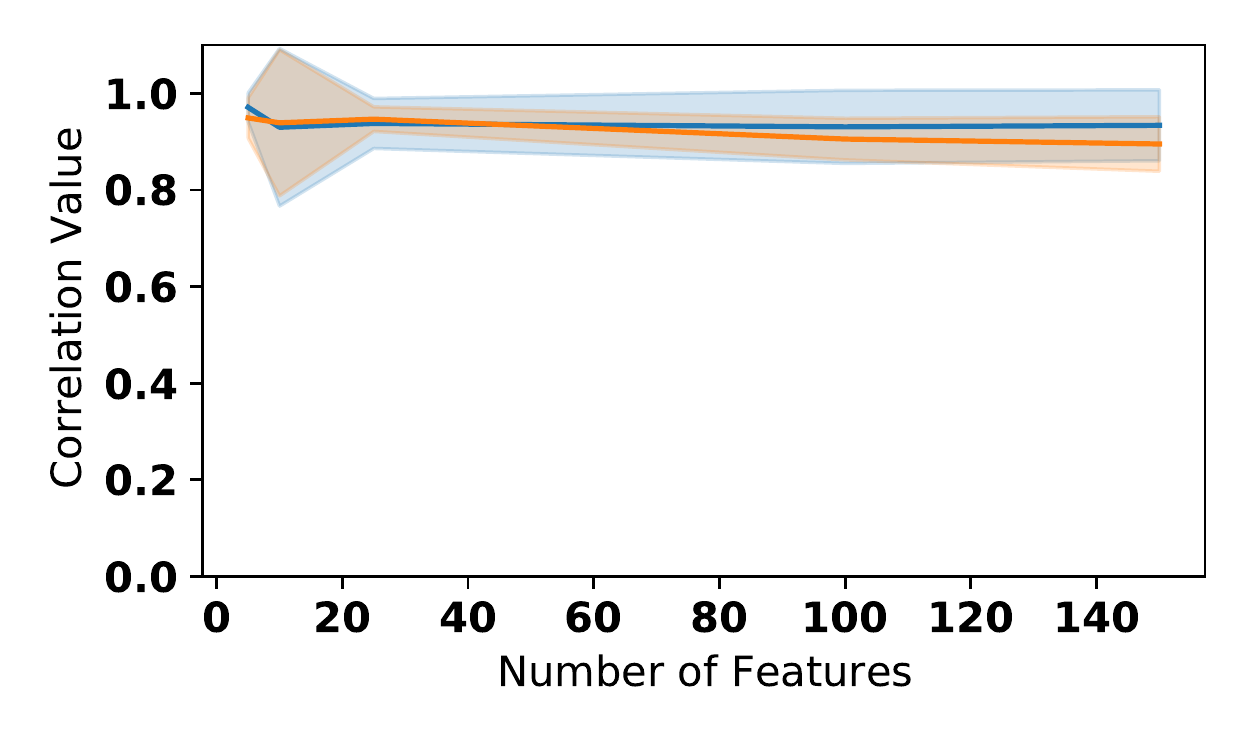}
    \includegraphics[width=0.31\linewidth]{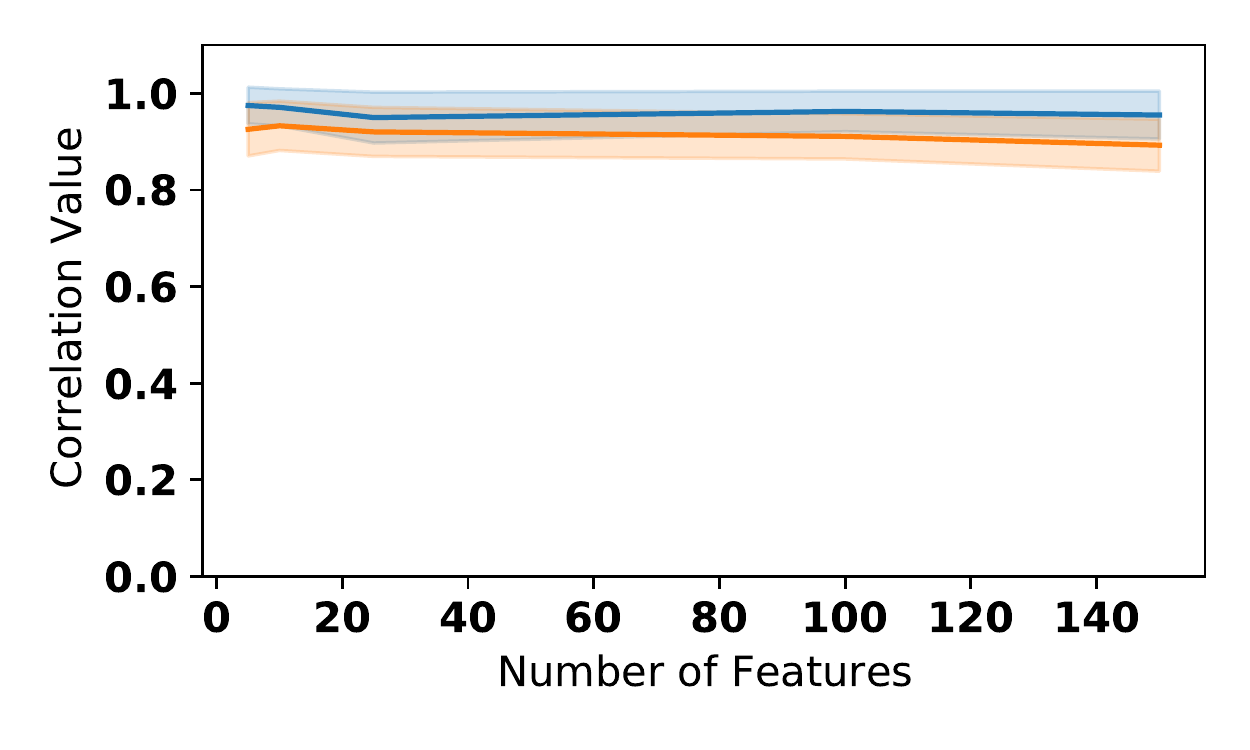}
  \caption{Correlation of predicted outputs in models trained on synthetic data with model perturbations across different number of features  (Blue: Pearson, Orange: Spearman correlation). The predicted output of perturbed models are still highly correlated to those without perturbation.}
  \label{fig:pred_corr}
\end{figure}

In this section we evaluate the stability of feature importances when inputs and models are perturbed. In all of the following experiment settings, the predicted outputs from the perturbed models and the original models are highly correlated (an example for model perturbation is shown in Figure \ref{fig:pred_corr} for synthetic data). This ensures that our models have very similar performance and the results are minimally affected by discrepancies between model predictions.

\vspace{-.3cm}
\subsubsection{Stability of Feature Importances When Inputs Are Perturbed}
\begin{figure}[h]
  \centering
  \scriptsize
 Model: \hspace{1cm} XGBoost \hspace{1.5cm} Gradient Boosting Machine \hspace{0.7cm} Random Forest \\
 \includegraphics[width=0.09\linewidth]{images/simulation/legend.png}
  \includegraphics[width=0.29\linewidth]{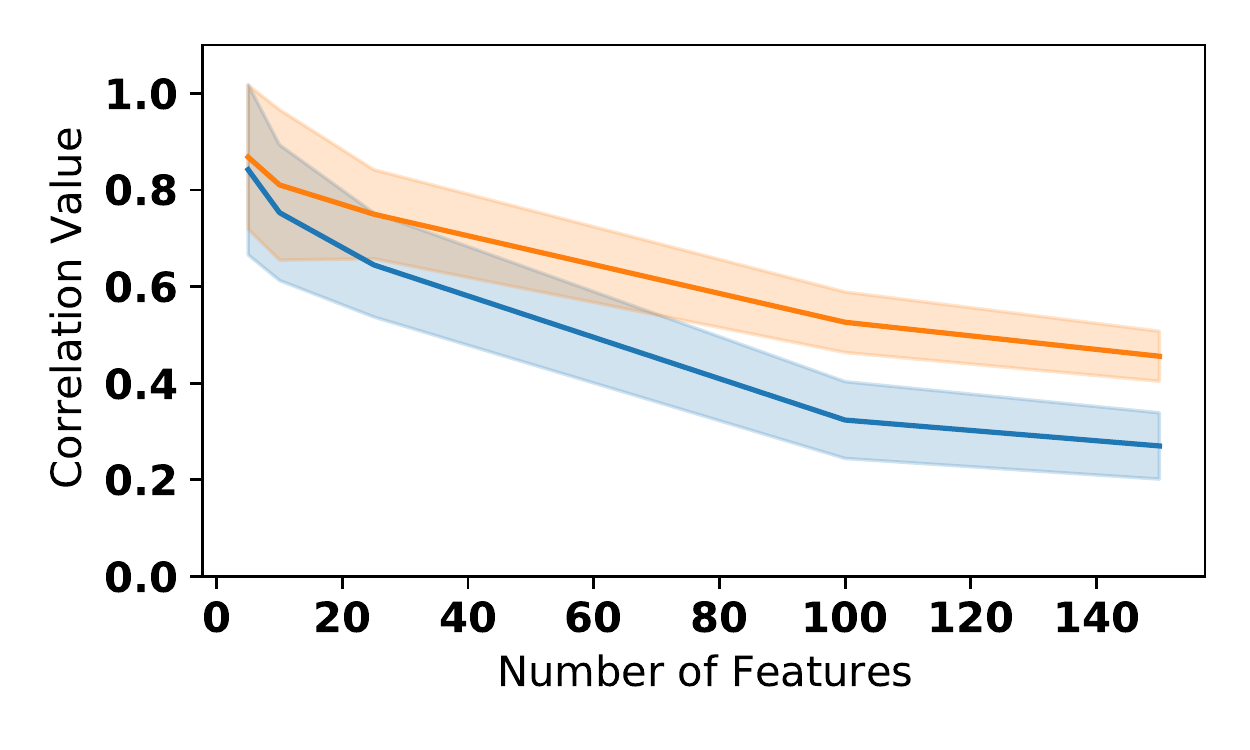}
  \includegraphics[width=0.29\linewidth]{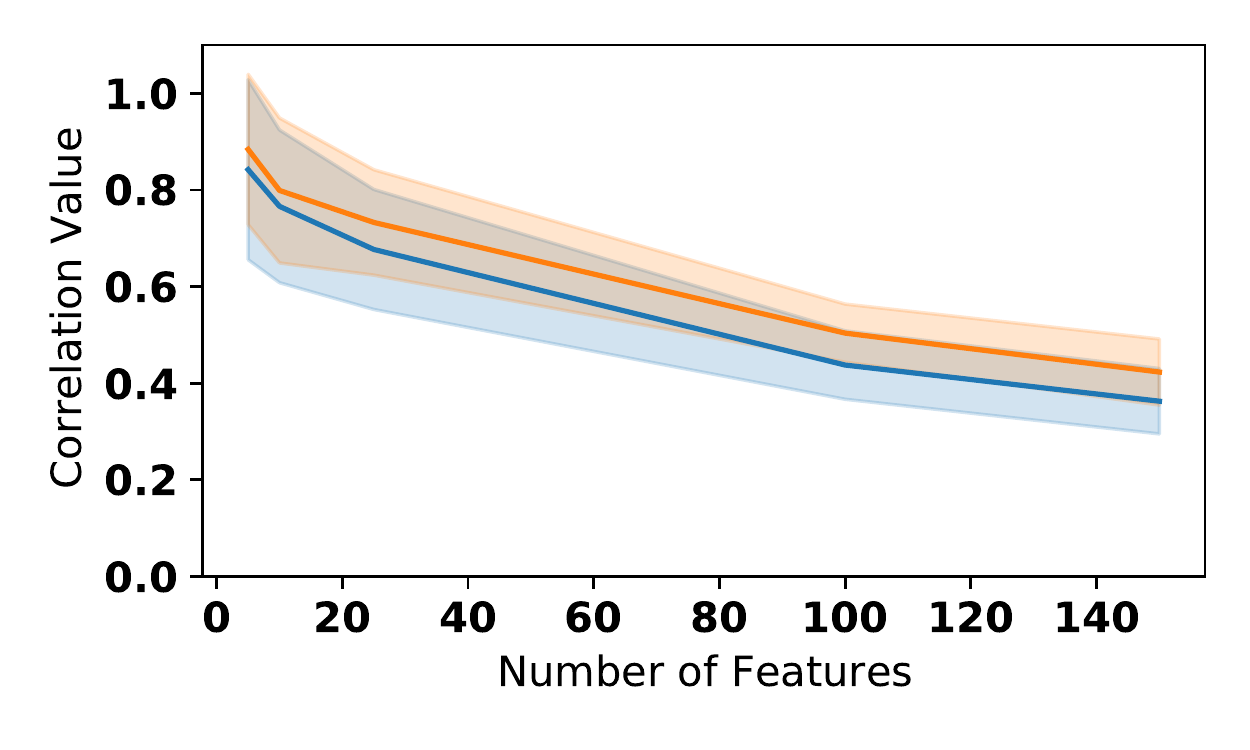}
  \includegraphics[width=0.29\linewidth]{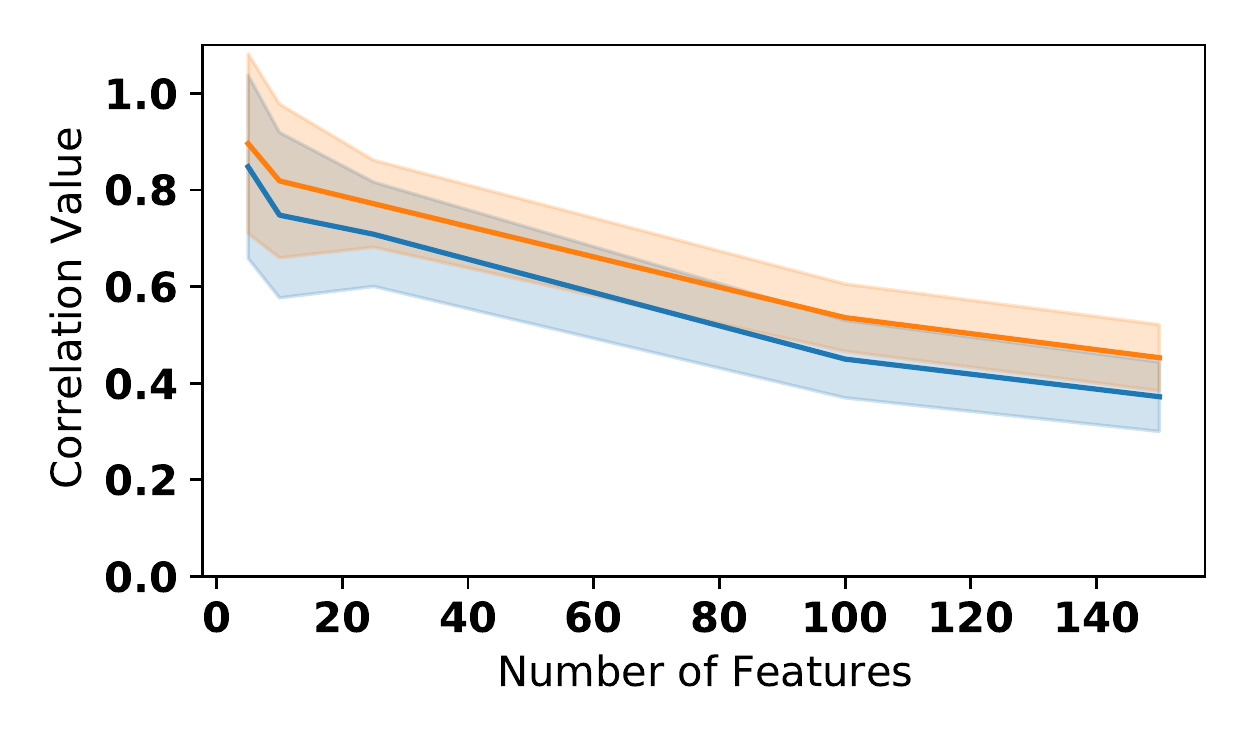}
  
   \caption{Correlation of feature importances (Blue: gain, Orange: SHAP) for models trained with low input perturbation on synthetic data. SHAP is more stable across all models although both SHAP and gain both suffer from lack of stability.}
  \label{fig:low_perturbed_input}
\end{figure}

\begin{figure*}[ht]
  \centering
  \scriptsize
    Model Type: XGBoost \\
        Low noise added to input \hspace{1cm} Medium noise \hspace{1.7cm} Large noise \\
    \hspace*{10ex}
    \includegraphics[width=0.29\linewidth]{images/simulation/all_noise0.5_XGB_NOHP_spearmancorr.pdf}
    \includegraphics[width=0.29\linewidth]{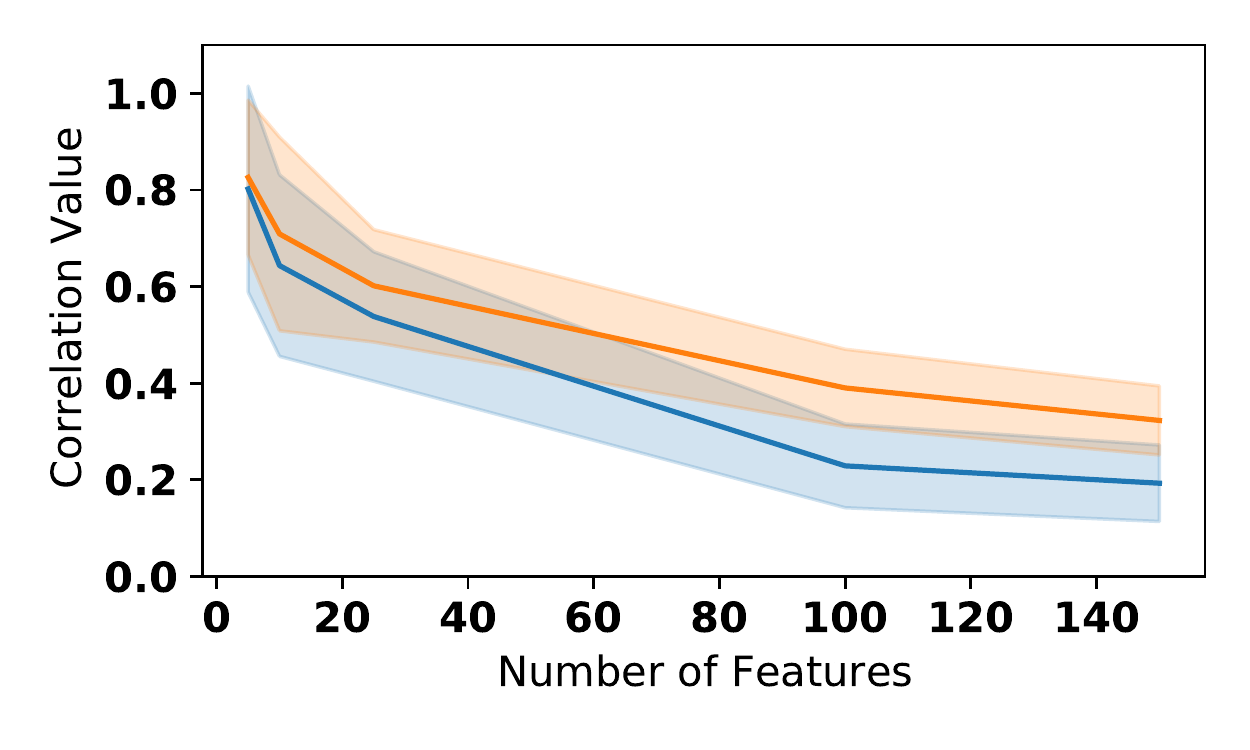}
    \includegraphics[width=0.29\linewidth]{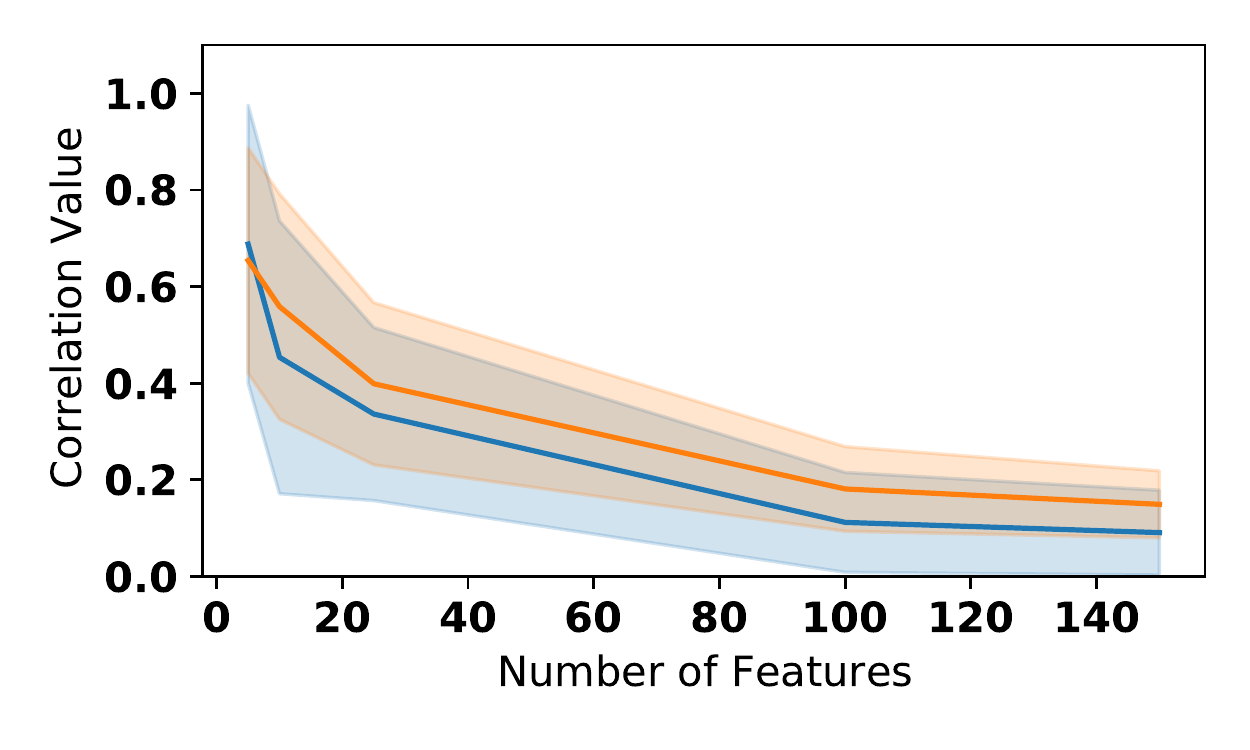}
    
    Model Type: Gradient Boosting \\
        Low noise added to input \hspace{1cm} Medium noise \hspace{1.7cm} Large noise \\
      \includegraphics[width=0.09\linewidth]{images/simulation/legend.png}
    \includegraphics[width=0.29\linewidth]{images/simulation/all_noise0.5_GB_NOHP_spearmancorr.pdf}
    \includegraphics[width=0.29\linewidth]{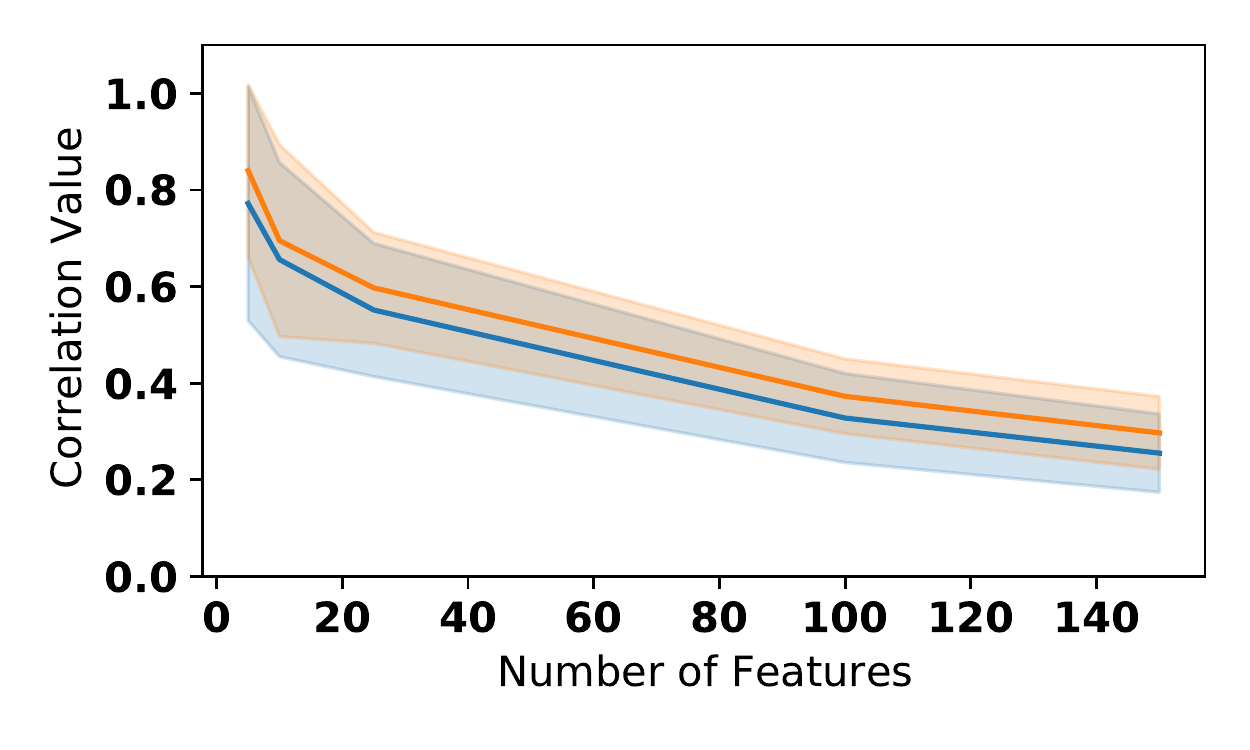}
    \includegraphics[width=0.29\linewidth]{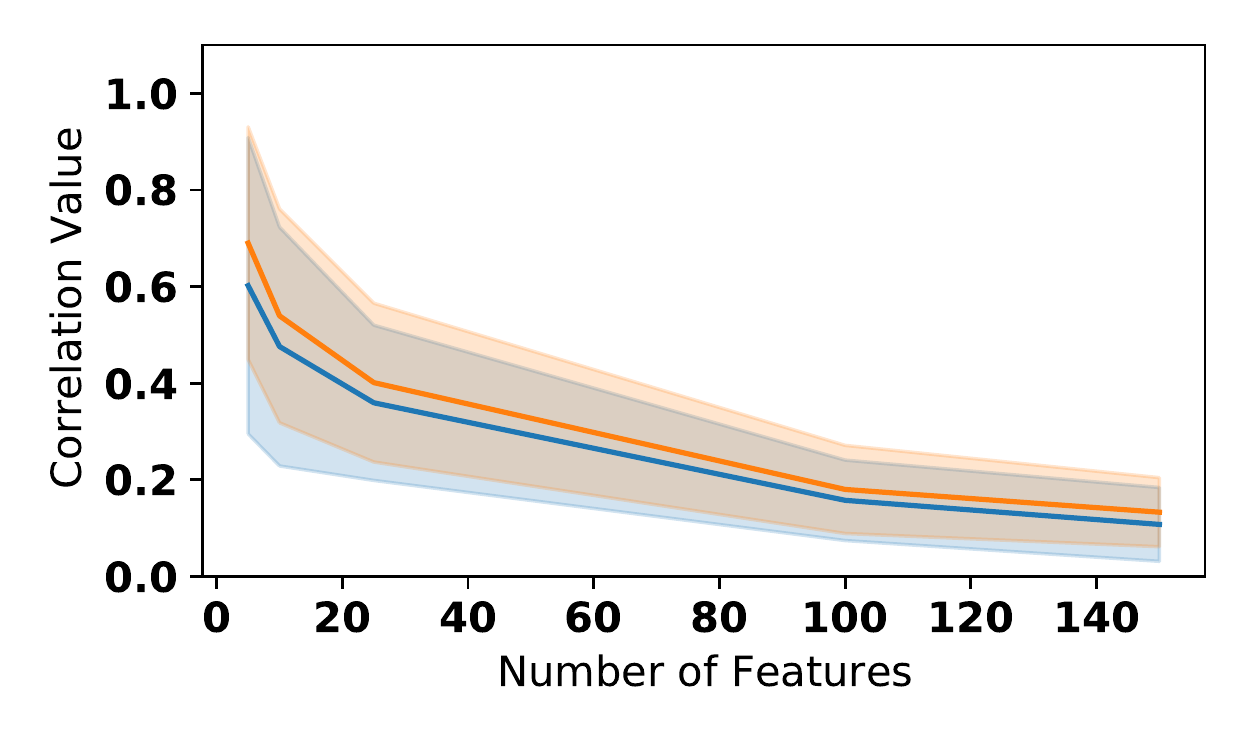}
    
    Model Type: Random Forest \\
    Low noise added to input \hspace{1cm} Medium noise \hspace{1.7cm} Large noise \\
    \hspace{10ex}
    \includegraphics[width=0.29\linewidth]{images/simulation/all_noise0.5_RF_NOHP_spearmancorr.pdf}
    \includegraphics[width=0.29\linewidth]{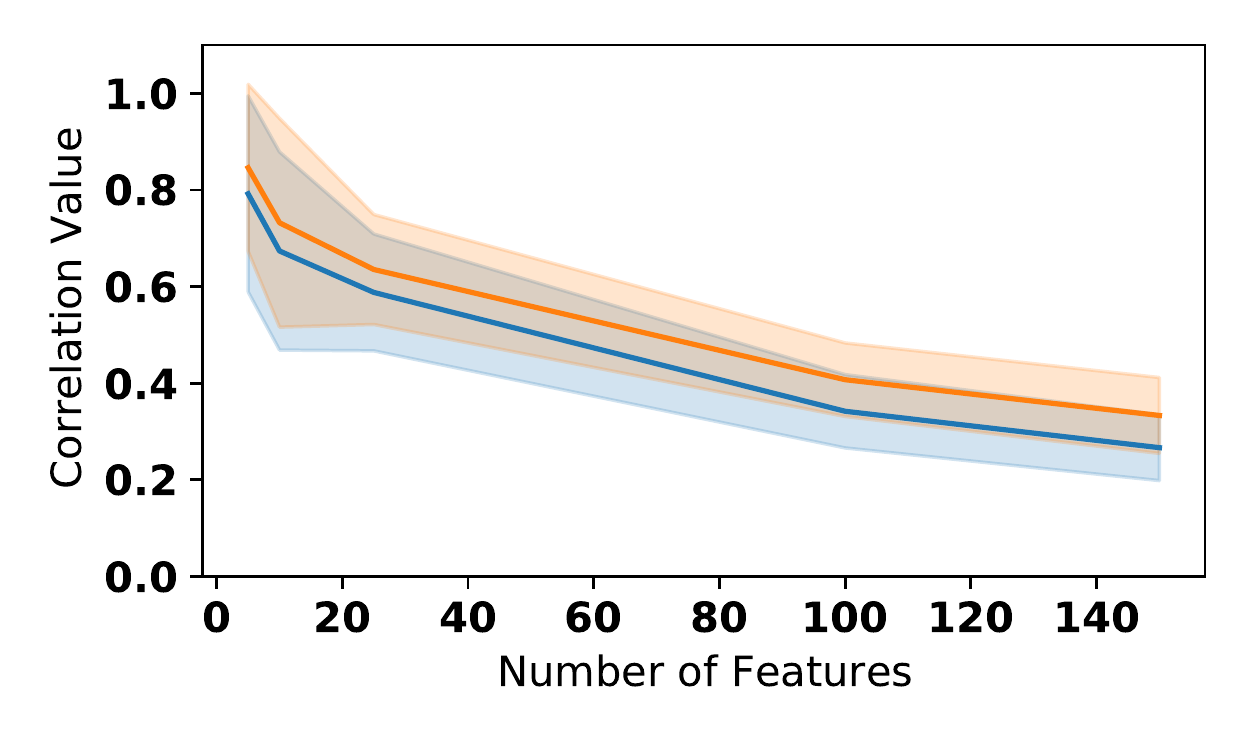}
    \includegraphics[width=0.29\linewidth]{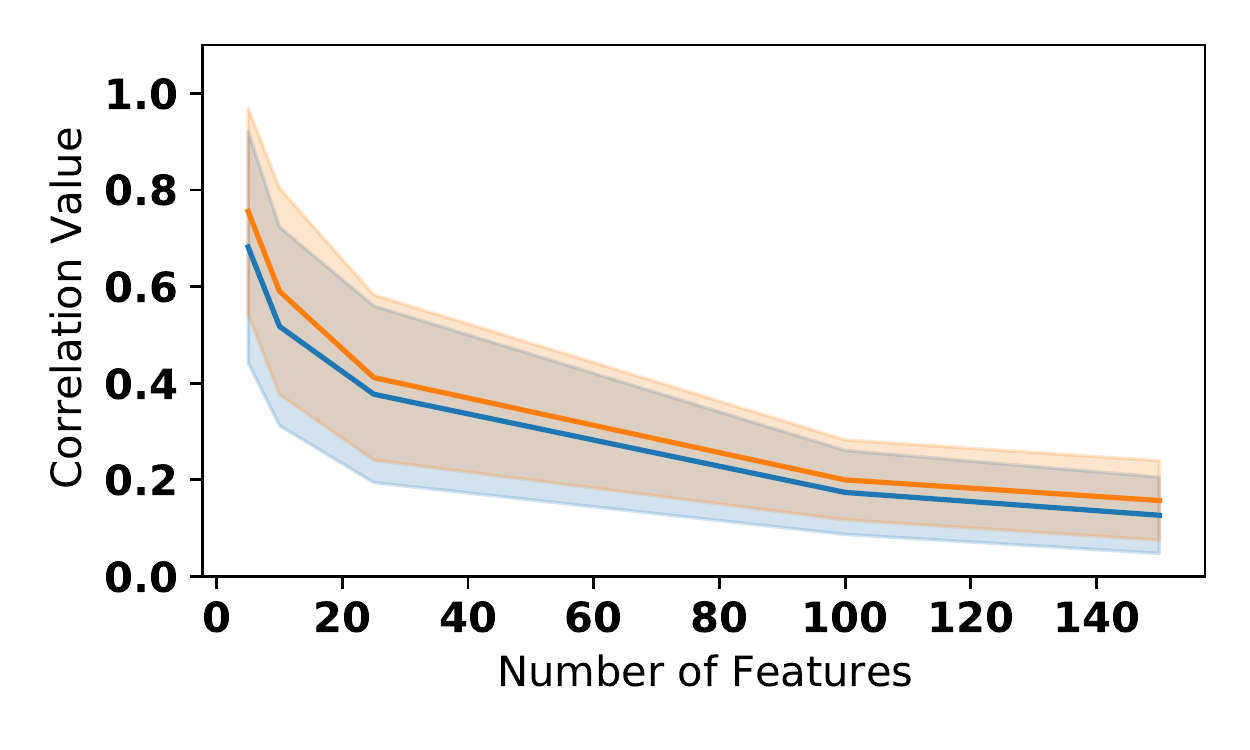}
  \caption{Correlation of feature importances (Blue: gain, Orange: SHAP) for models trained with input perturbation on synthetic data. SHAP is slightly more stable than gain at low level of noise but are comparable as noise increases.}
  \label{fig:all_perturbed_input}
\end{figure*}

Figure \ref{fig:low_perturbed_input} shows us a glimpse of this analysis for low level of noise on synthetic data. From this figure, we see that SHAP is more stable than gain feature importances when we add a small noise to the perturbed input, especially for XGBoost. This uplift between gain and SHAP, however, decreases as noise increases across all models as shown in Figure \ref{fig:all_perturbed_input}. We can also see from Figure \ref{fig:all_perturbed_input} that unsurprisingly stability decreases as the level of noise and the number of features increase. 


\begin{figure*}[h]
  \centering
  \scriptsize
    Dataset: Forest Fire (\# Features: 12)\\
    XGBoost \hspace{1.7cm} Gradient Boosting \hspace{1cm} Random Forest \\

  \includegraphics[width=0.31\linewidth]{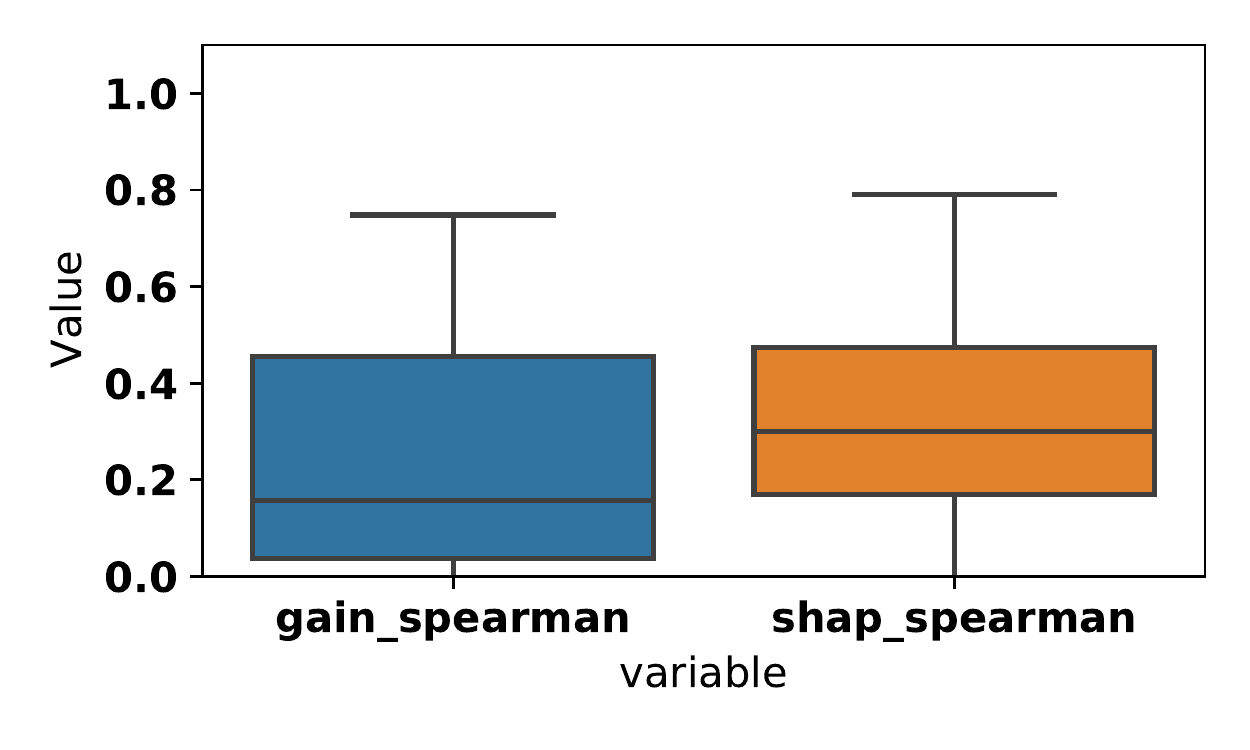}
  \includegraphics[width=0.31\linewidth]{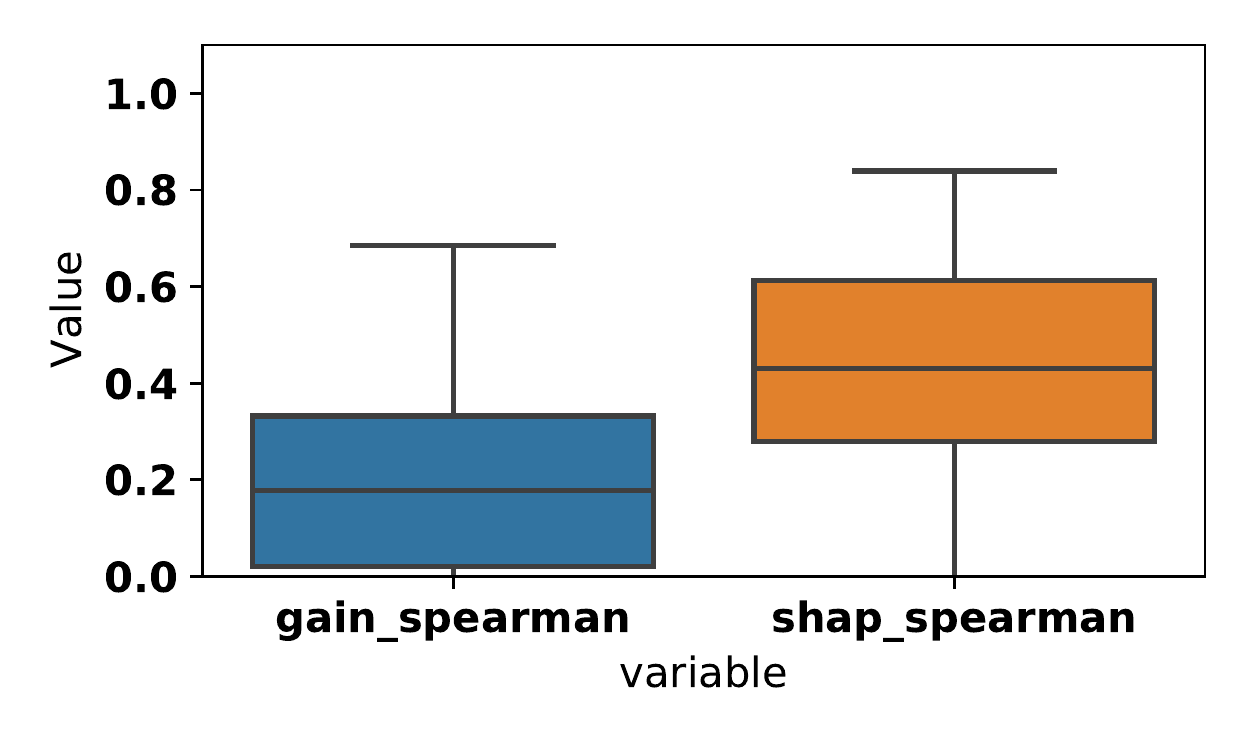}
  \includegraphics[width=0.31\linewidth]{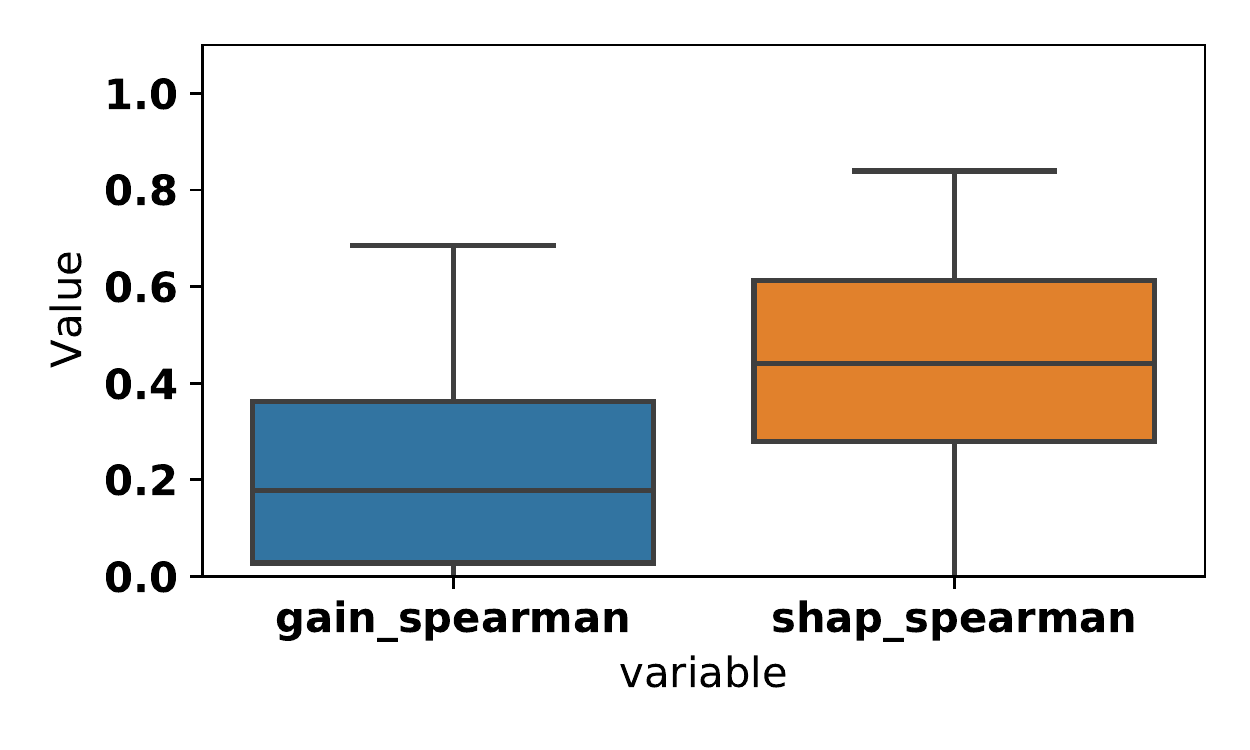}
  
  Dataset: Concrete (\# Features: 8)\\
    XGBoost \hspace{1.7cm} Gradient Boosting \hspace{1cm} Random Forest \\

  \includegraphics[width=0.31\linewidth]{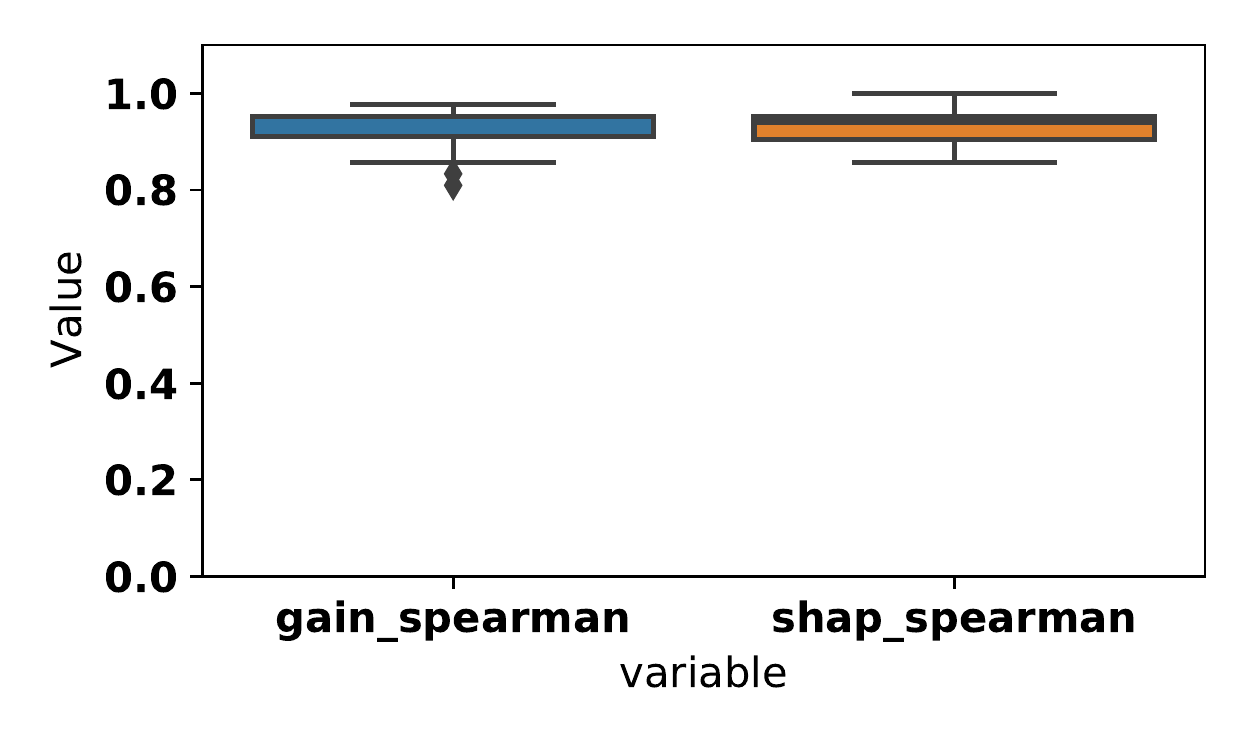}
  \includegraphics[width=0.31\linewidth]{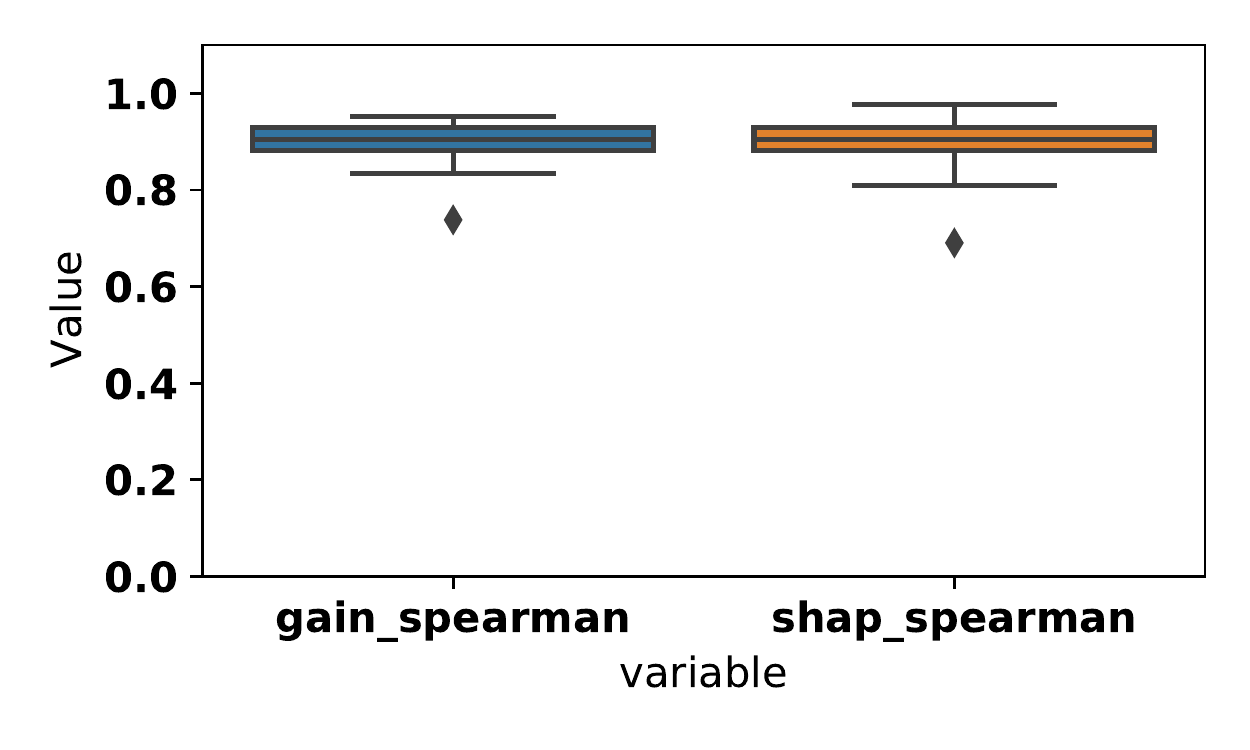}
  \includegraphics[width=0.31\linewidth]{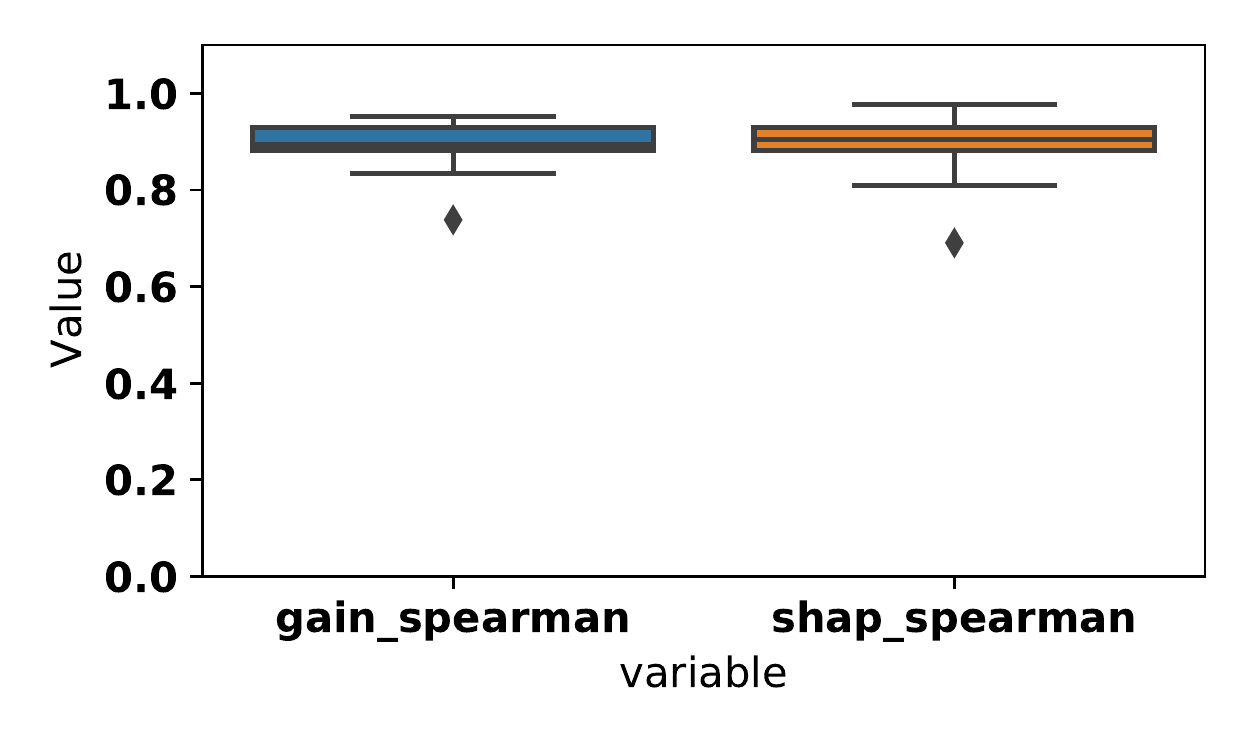}
  
  Dataset: Auto MPG (\# Features: 7)\\
    XGBoost \hspace{1.7cm} Gradient Boosting \hspace{1cm} Random Forest \\

  \includegraphics[width=0.31\linewidth]{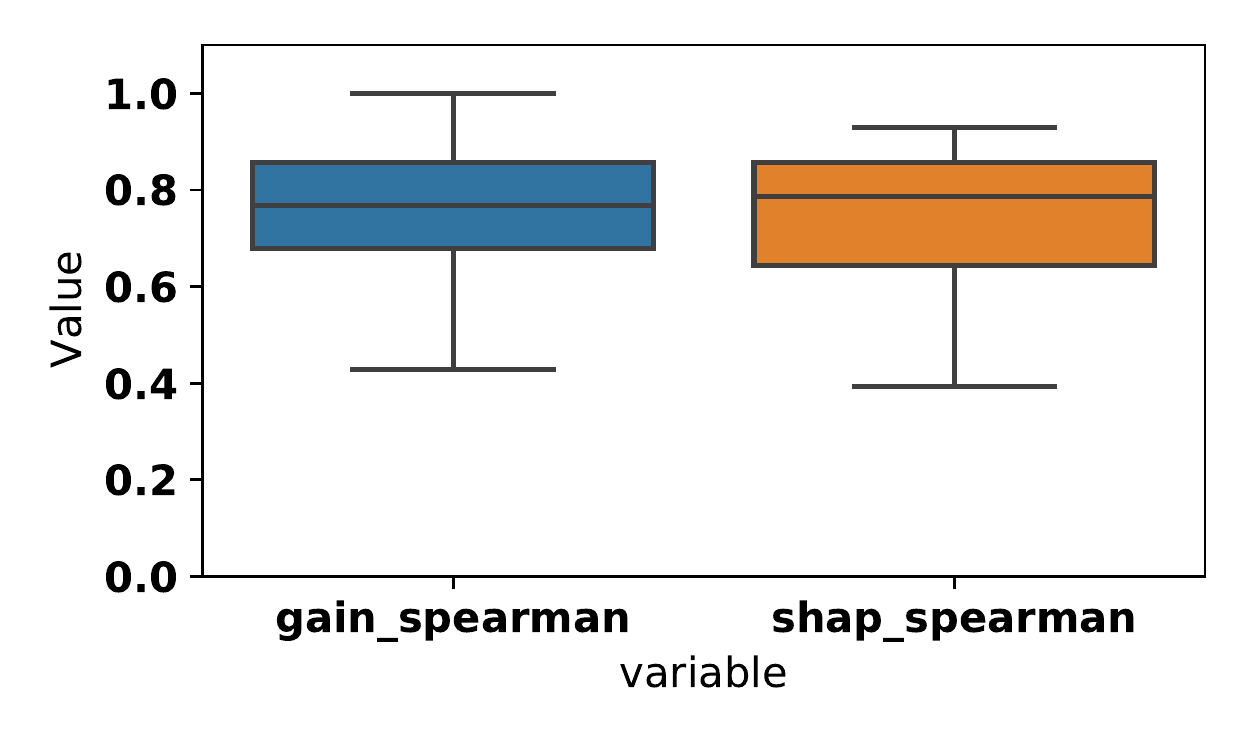}
  \includegraphics[width=0.31\linewidth]{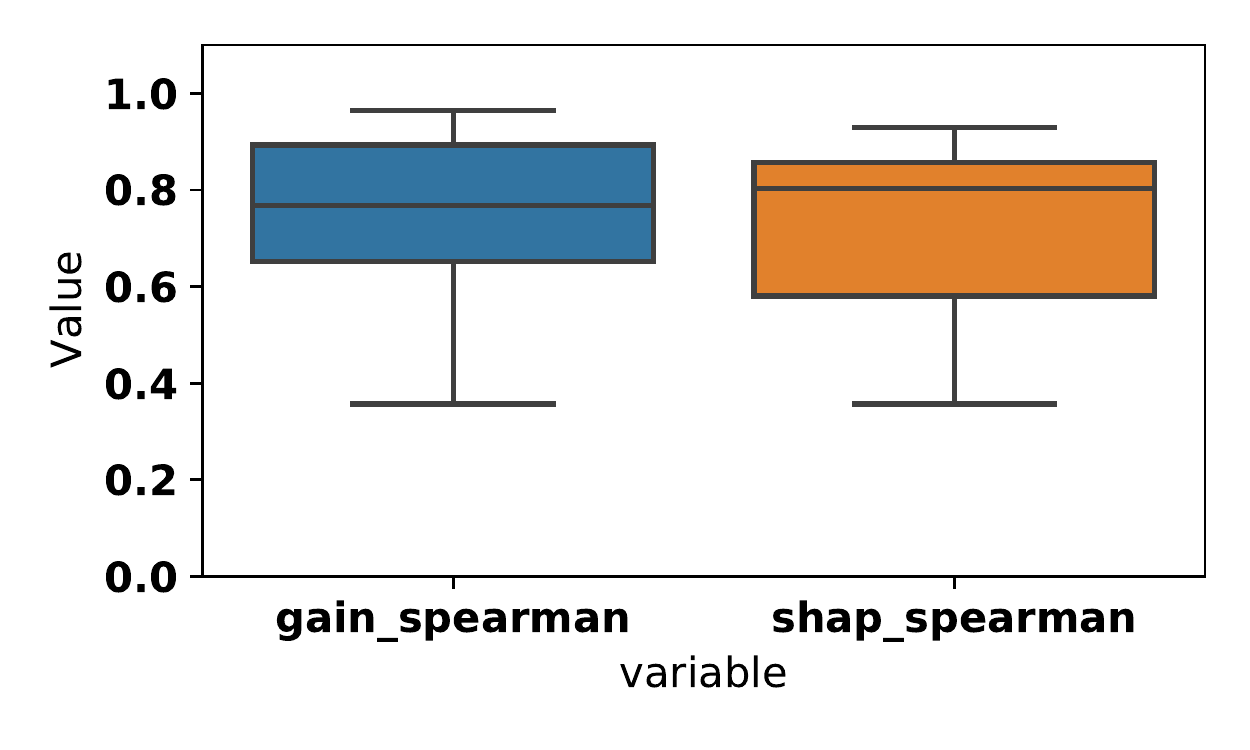}
  \includegraphics[width=0.31\linewidth]{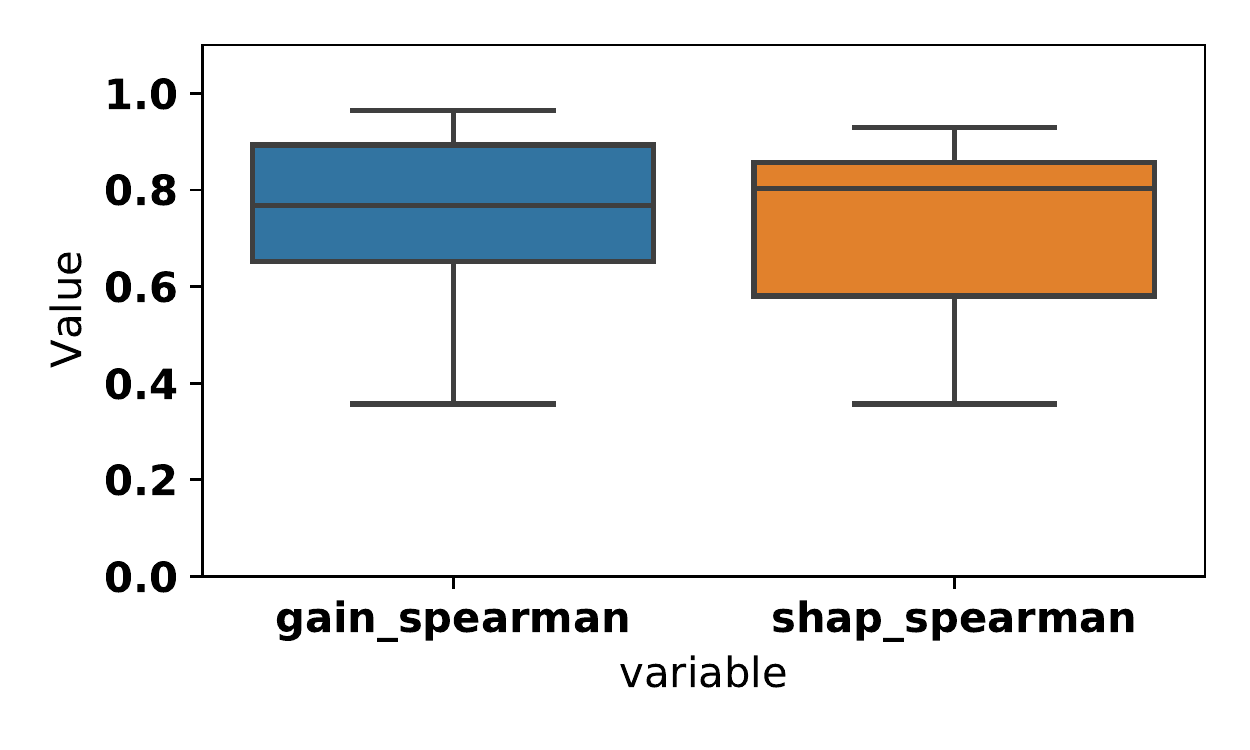}
  
  Dataset: Company Finance (\# Features: 892)\\
     XGBoost \hspace{1.7cm} Gradient Boosting \hspace{1cm} Random Forest \\

  \includegraphics[width=0.31\linewidth]{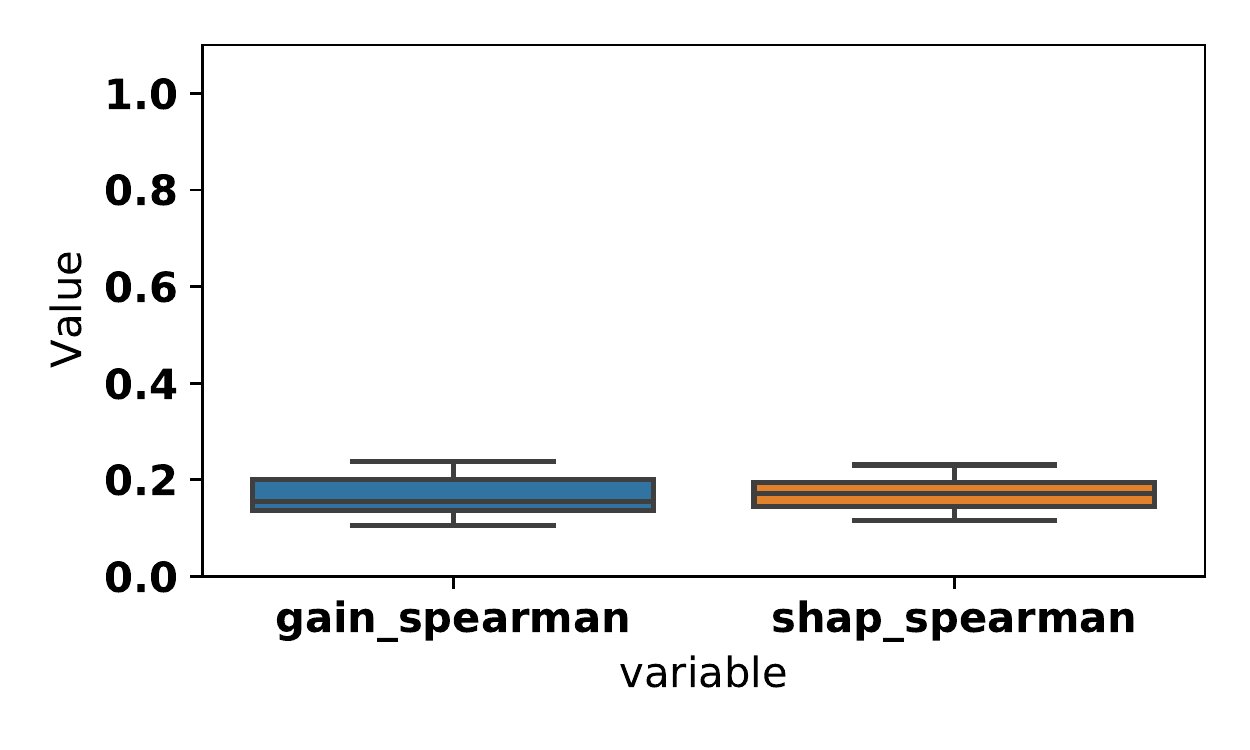}
  \includegraphics[width=0.31\linewidth]{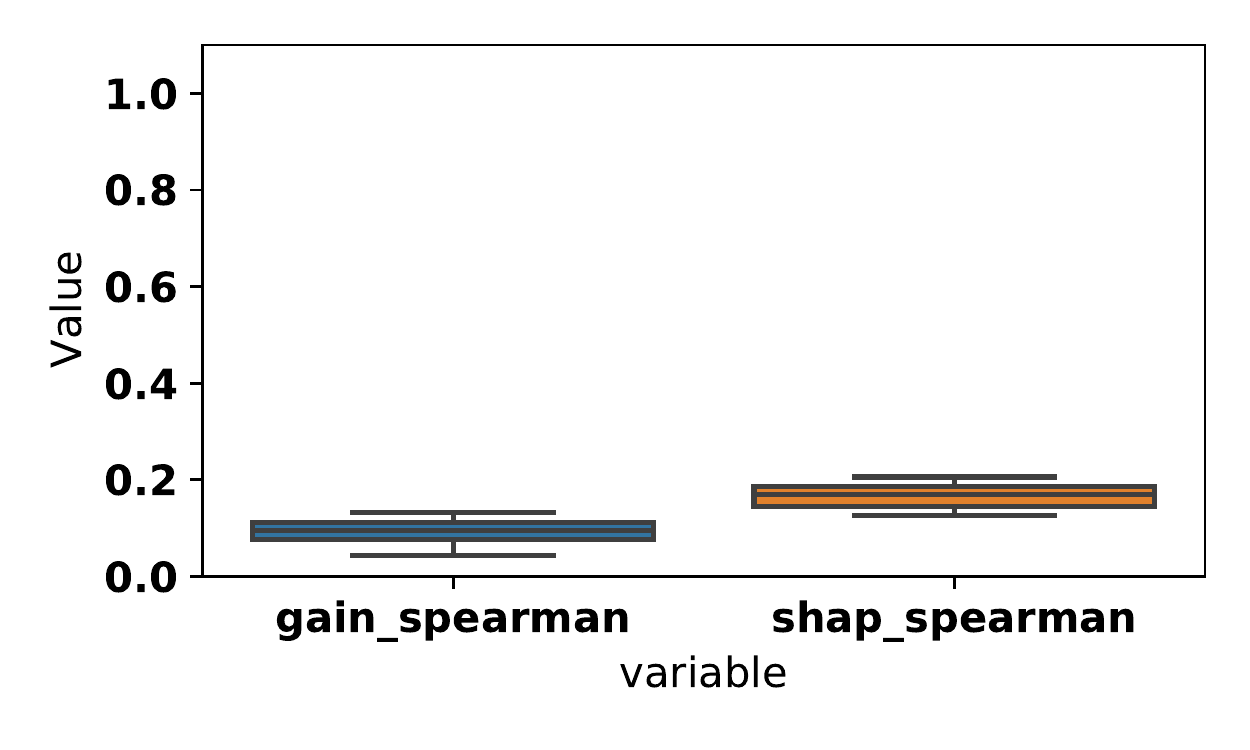}
  \includegraphics[width=0.31\linewidth]{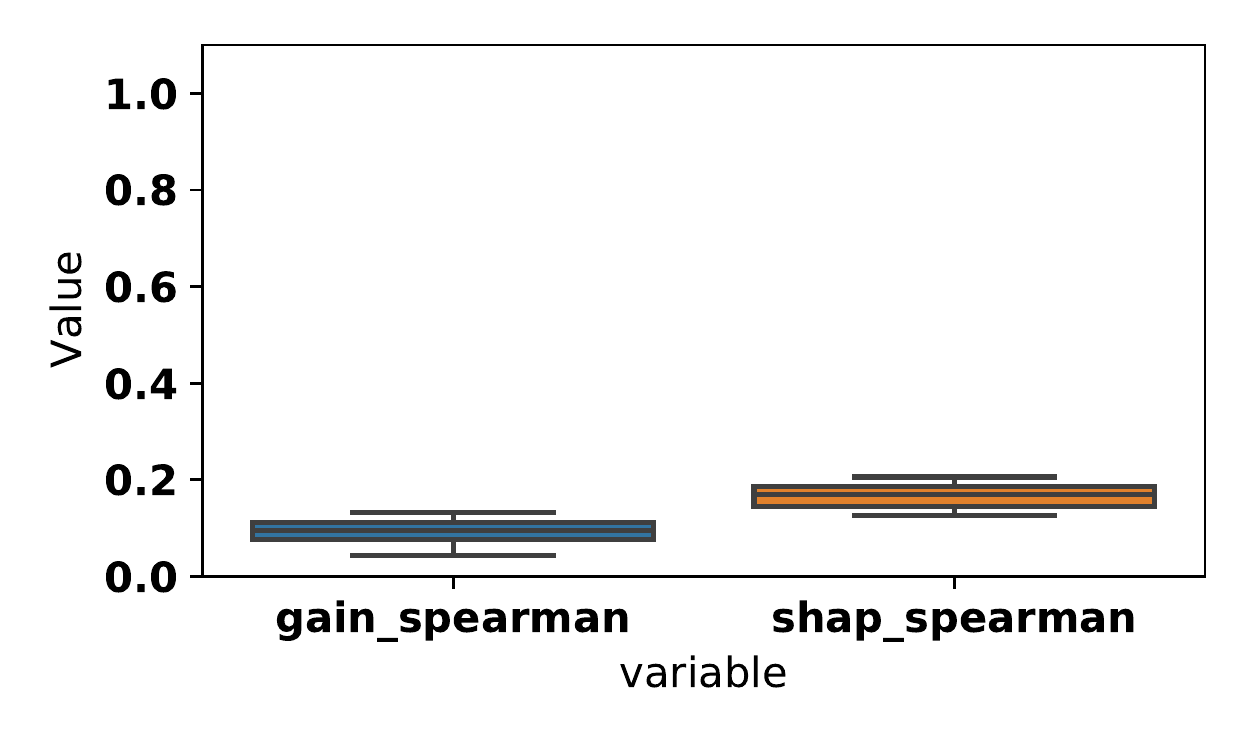}
  \caption{Correlation of feature importances (Blue: gain, Orange: SHAP) for models trained with input perturbations (low noise) on real-world datasets.  SHAP and gain both lack stability overall although SHAP is slightly more stable for certain datasets. }
\label{fig:real_world_noisy_inputs}
\end{figure*}

As shown in Figure \ref{fig:real_world_noisy_inputs}, we see that in real-world datasets when a low noise is injected to the input, the correlations of gain and SHAP feature importances drop very low. For example, in Forest Fire dataset, feature importances correlation averages to around 50\% for SHAP while it averages to around 20\% for gain. In Company Finance dataset, both gain and SHAP has either 20\% correlation or lower. We discover that SHAP is slightly more stable than gain for Forest Fire and Company Finance as can be seen on Figure \ref{fig:real_world_noisy_inputs}, although this is not consistent across all datasets. We also observe low correlations with increasing level of noise.  


\vspace{-.3cm}
\subsubsection{Stability of Feature Importances When Models Are Perturbed}
\begin{figure*}[h]
 \centering
 \scriptsize
 Perturbation: Random seeds \\
 Model: \hspace{1cm} XGBoost \hspace{1.5cm} Gradient Boosting Machine \hspace{0.7cm} Random Forest \\ 
  \includegraphics[width=0.09\linewidth]{images/simulation/legend.png}
 \includegraphics[width=0.29\linewidth]{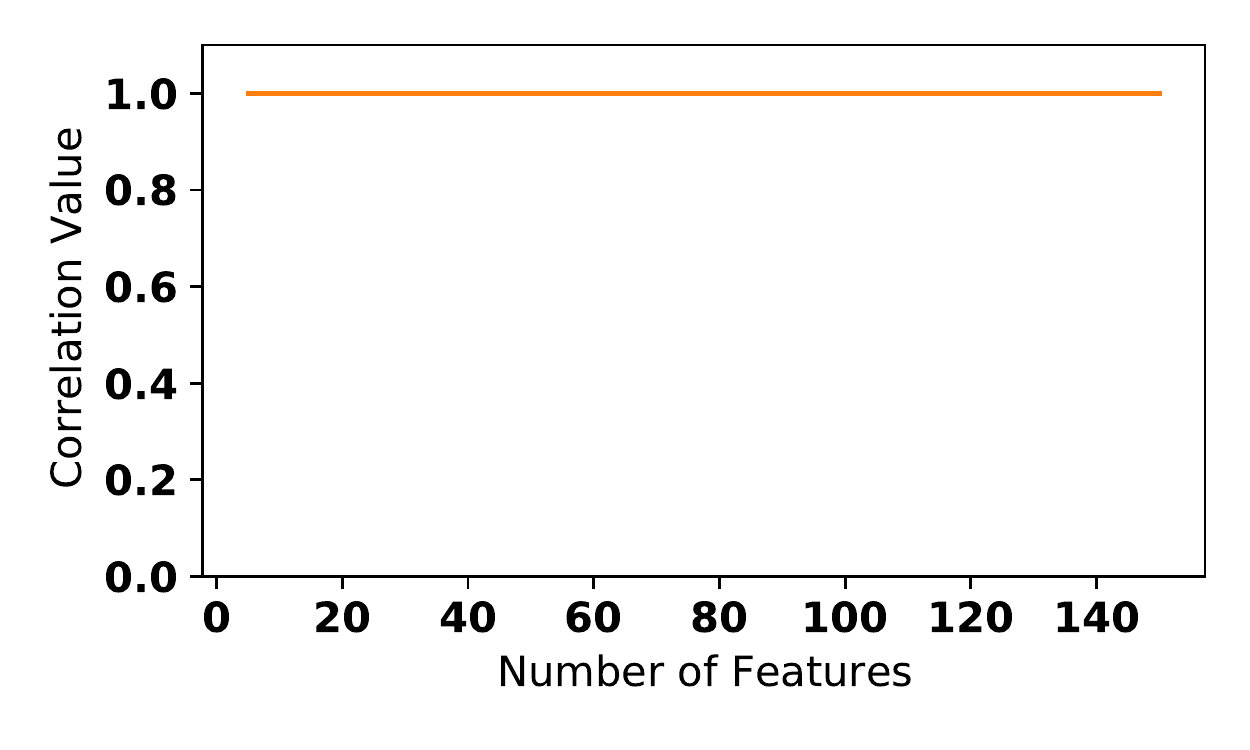}
 \includegraphics[width=0.29\linewidth]{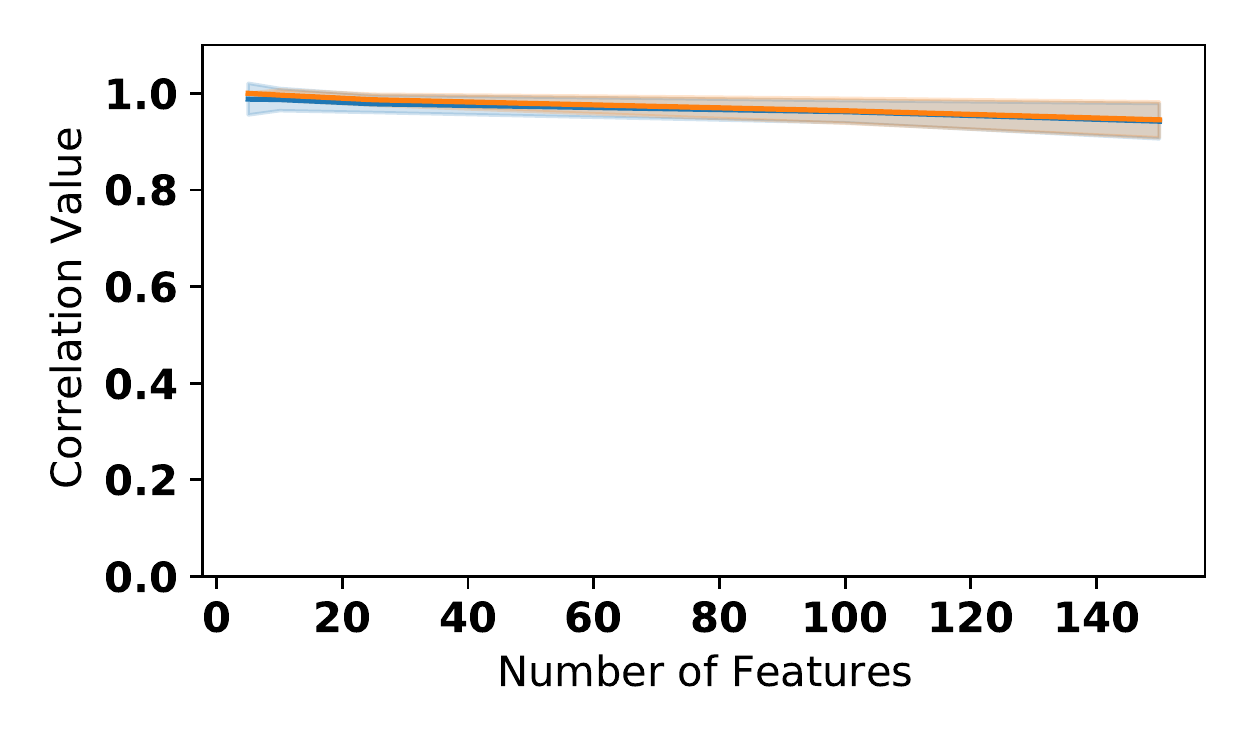}
 \includegraphics[width=0.29\linewidth]{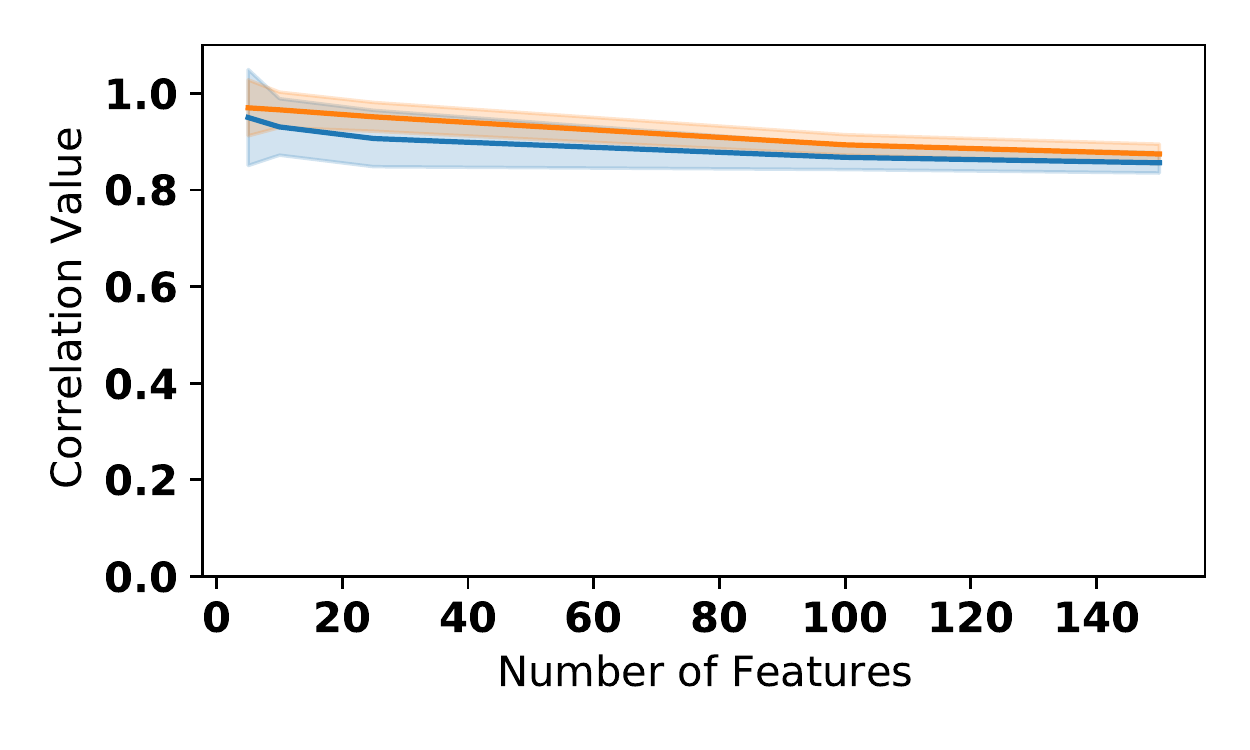} \\
 
 Perturbation: Hyperparameter settings\\
 Model: \hspace{1cm} XGBoost \hspace{1.5cm} Gradient Boosting Machine \hspace{0.7cm} Random Forest \\ 
 \hspace*{10ex}
 \includegraphics[width=0.29\linewidth]{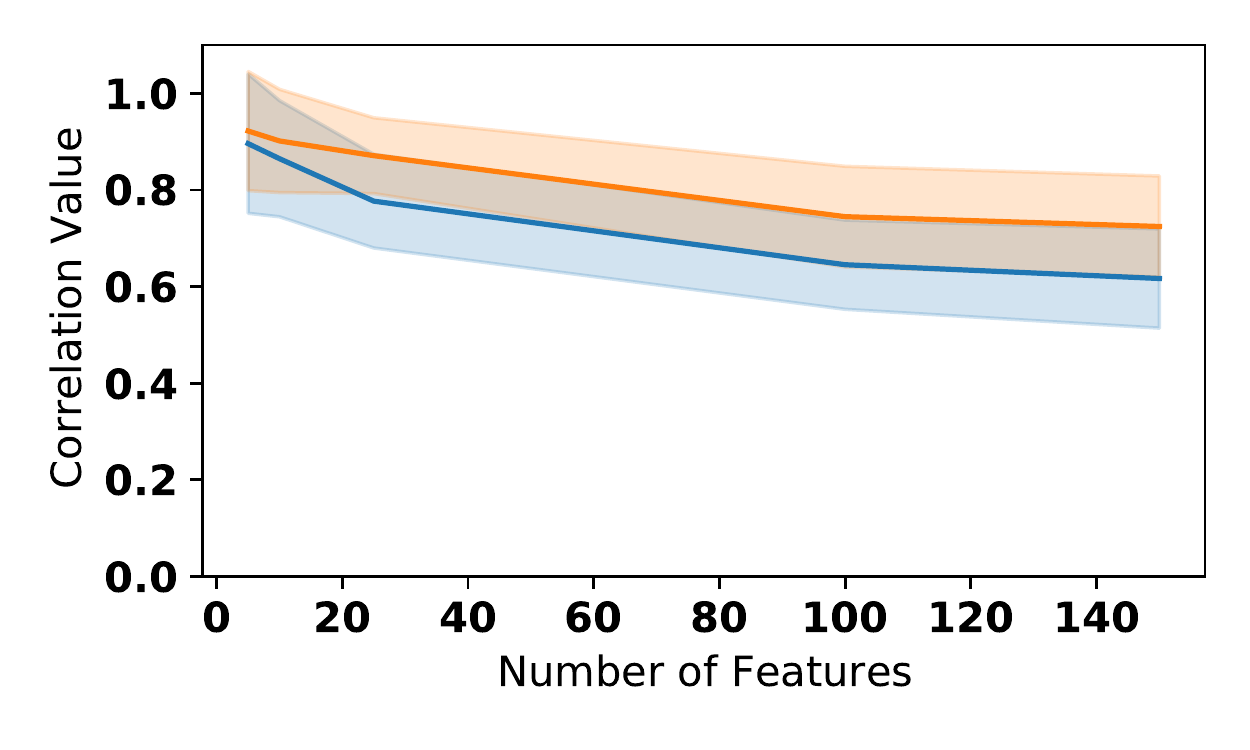}
 \includegraphics[width=0.29\linewidth]{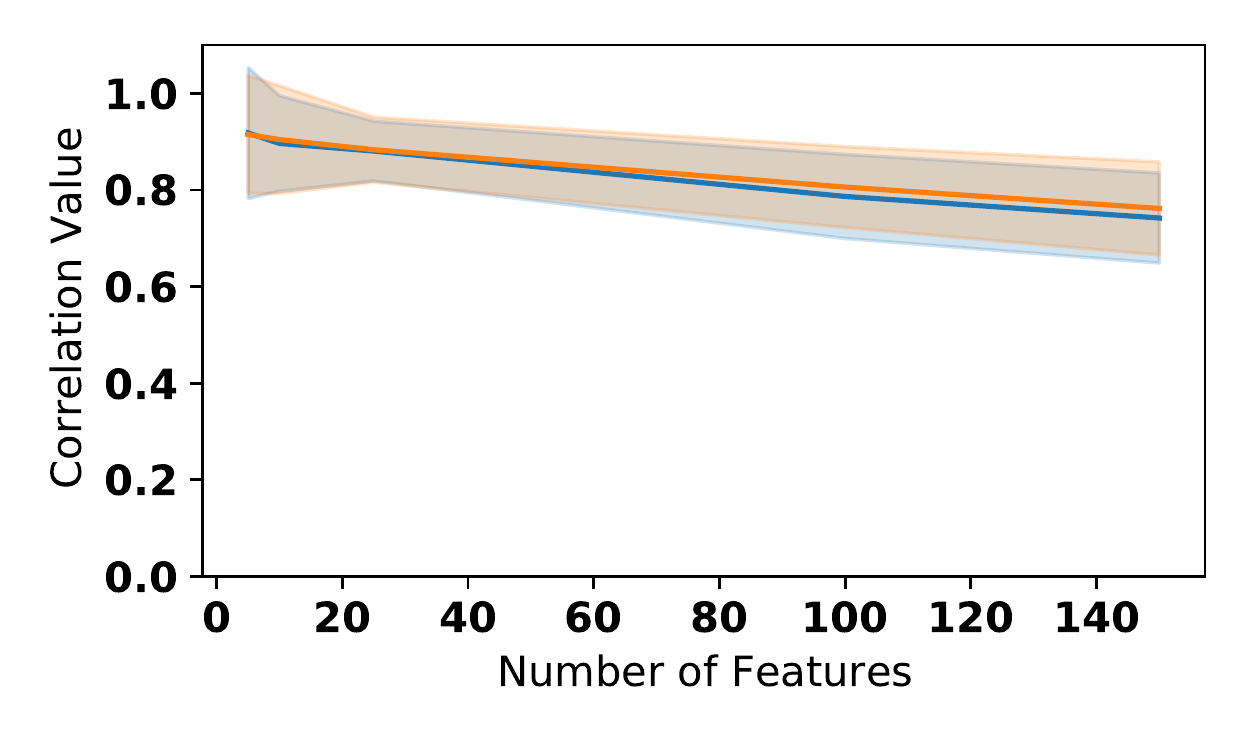}
 \includegraphics[width=0.29\linewidth]{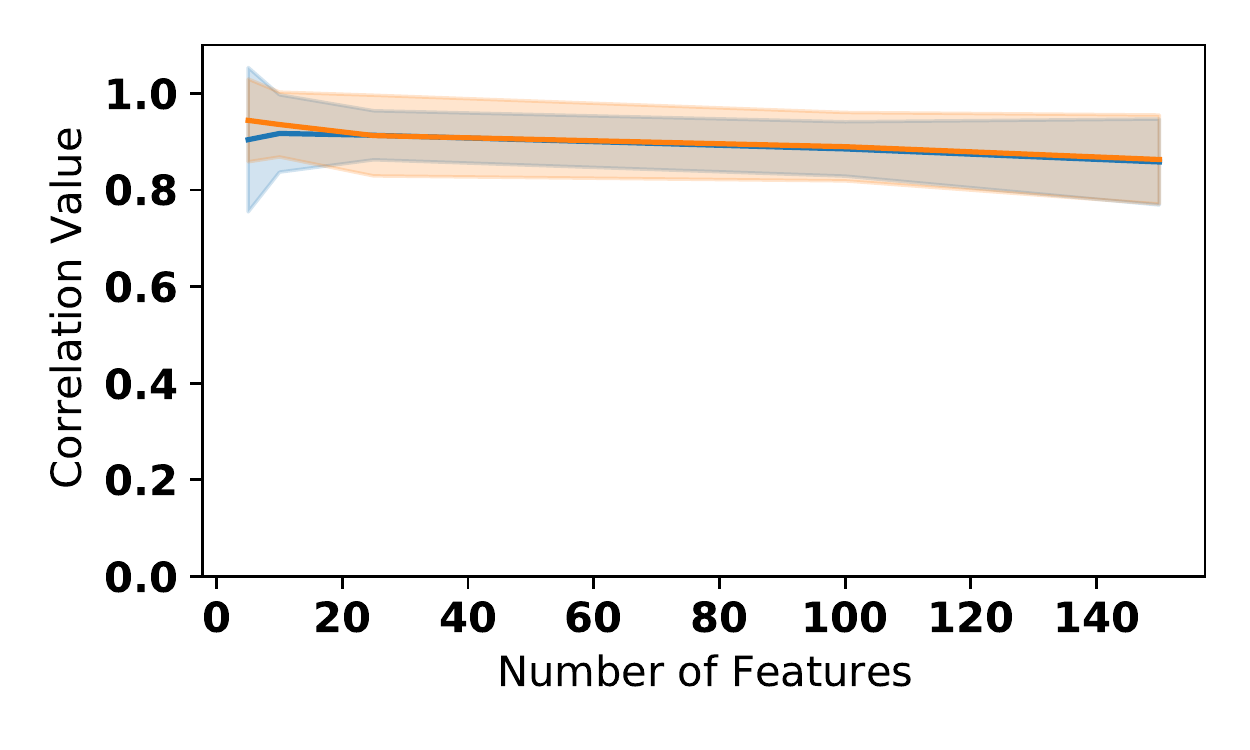} \\
  \caption{Correlation of feature importances (Blue: gain, Orange: SHAP) for models trained on synthetic data with model perturbations across different number of features.}
  \label{fig:perturb_model}
\end{figure*}

Figure \ref{fig:perturb_model} shows the correlation of feature importances when models are perturbed by initializing to a different random seed or by training with different hyperparameter settings. From this figure, we see that the correlation of feature importances is not greatly affected when models are perturbed for small number of features, but it drops significantly (to 80\% Spearman correlation for XGBoost and gradient boosting models) as the number of features increases to 150. We find that the correlation of SHAP feature importances is significantly higher compared to gain feature importances, especially in XGBoost trained with different hyperparameter. Although, for gradient boosting machine and random forest, we do not see the same uplift on stability for SHAP. Both gain and SHAP are equally stable for these models.

Moreover, we notice a strangely perfect correlation when training XGBoost without hyperparameter optimization but with different random seeds (See Figure \ref{fig:perturb_model}, top left). After further investigation, we discover that XGBoost is more deterministic when choosing features even when initialized with different random seeds. The results of our findings are expanded further in Appendix \ref{appendix:xgboost_deterministic}.  

\begin{figure}[h]
    \centering
    \scriptsize
    Dataset: Forest Fire (\# Features: 12) \\
     XGBoost  \hspace{1.7cm} Gradient Boosting Machine \hspace{1cm} Random Forest \\ 
    \includegraphics[width=0.31\linewidth]{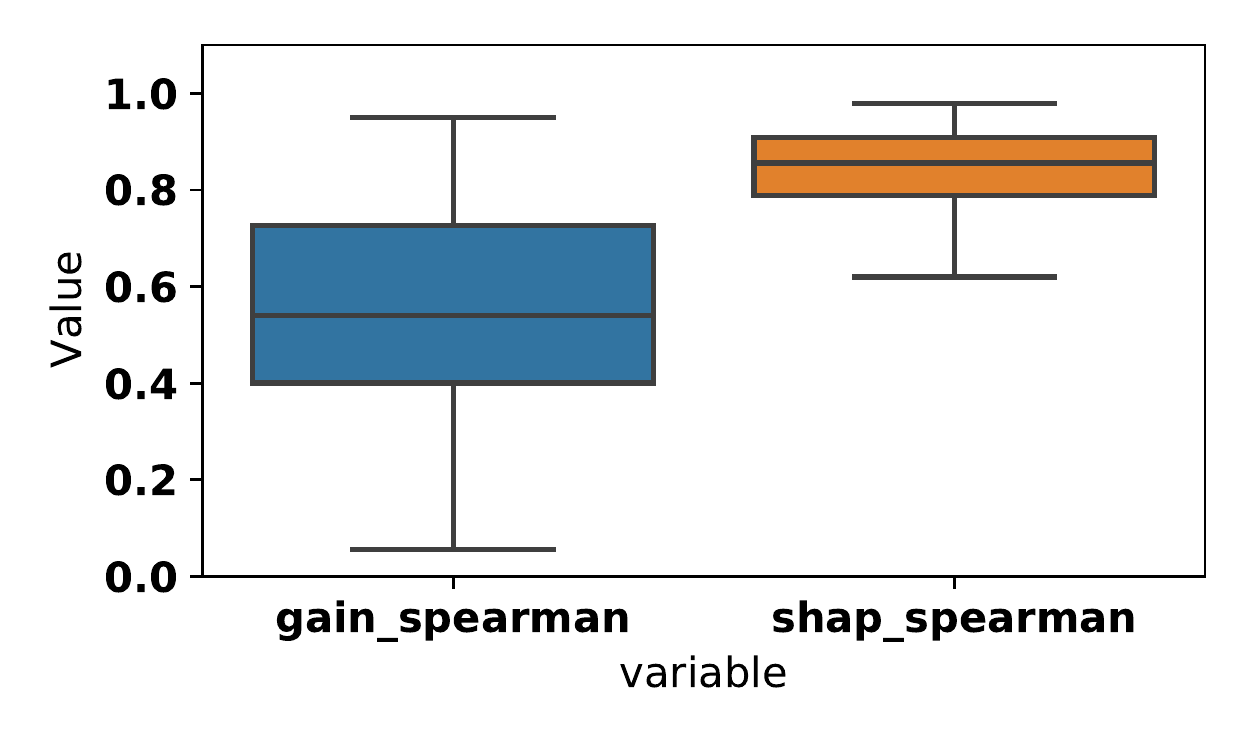}
    \includegraphics[width=0.31\linewidth]{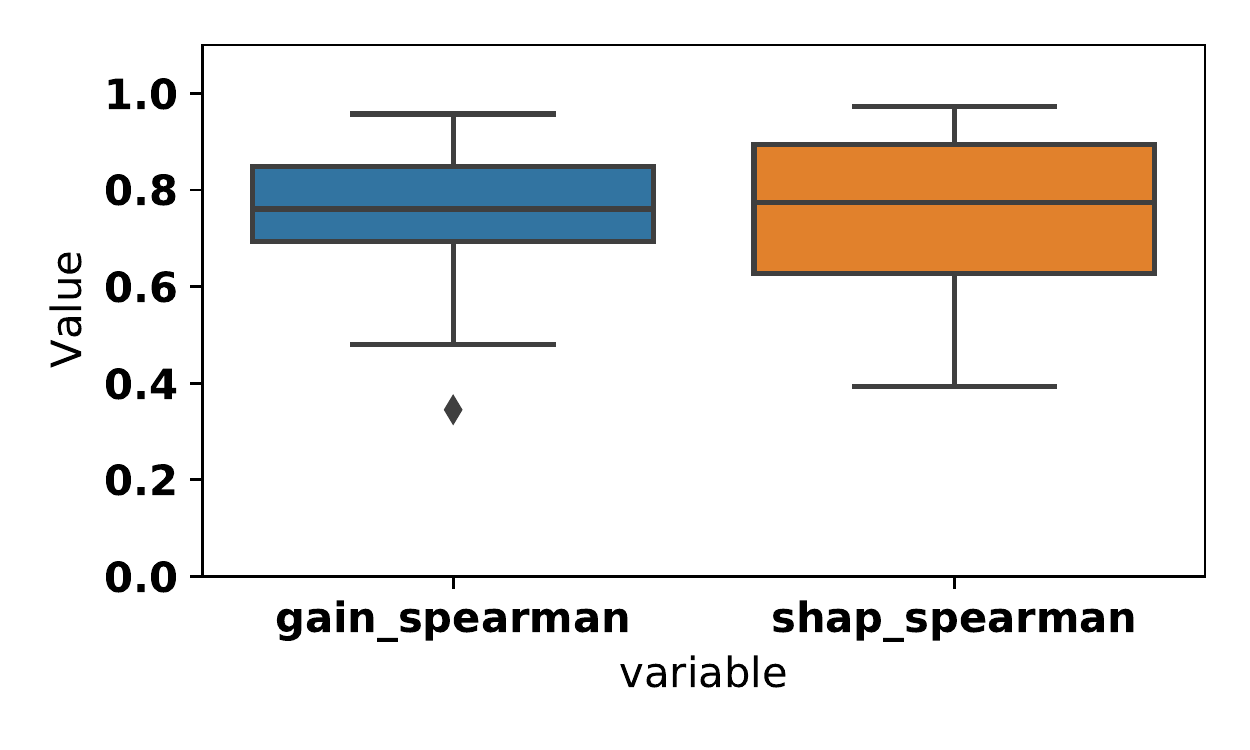}
    \includegraphics[width=0.31\linewidth]{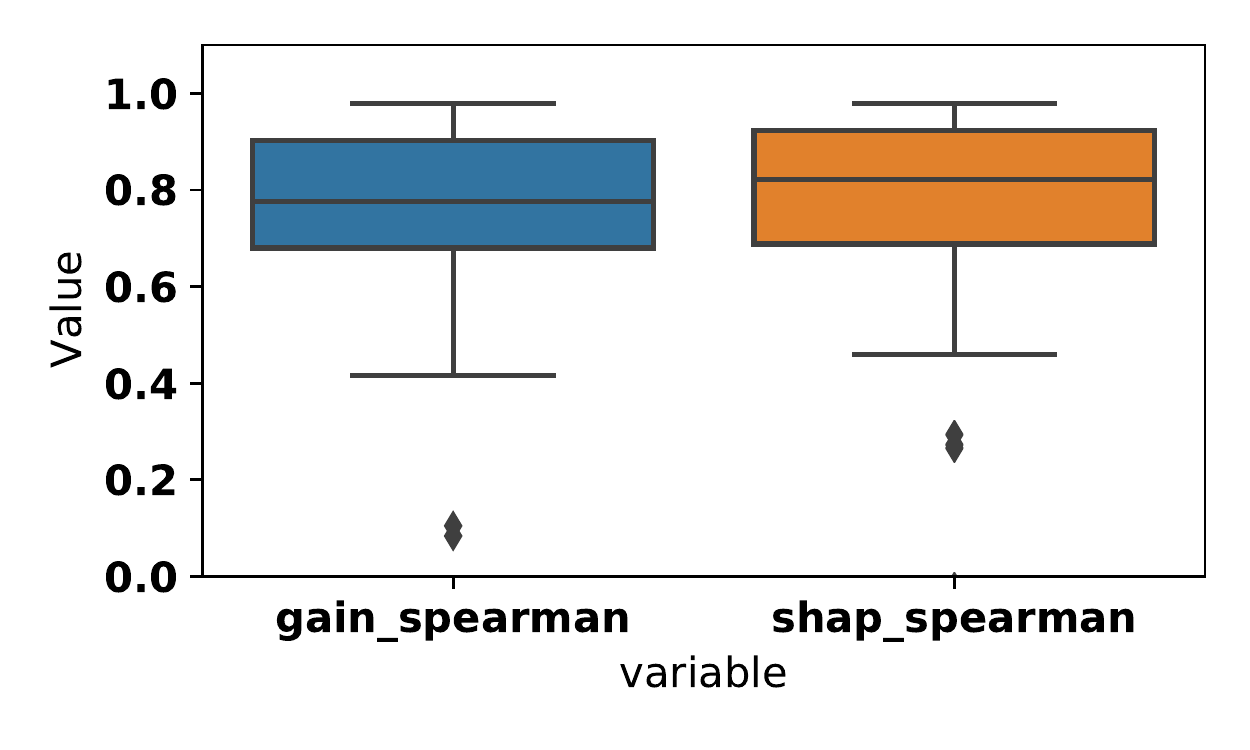}
    
    Dataset: Concrete (\# Features: 8) \\
     XGBoost  \hspace{1.7cm} Gradient Boosting Machine \hspace{1cm} Random Forest \\ 
    \includegraphics[width=0.31\linewidth]{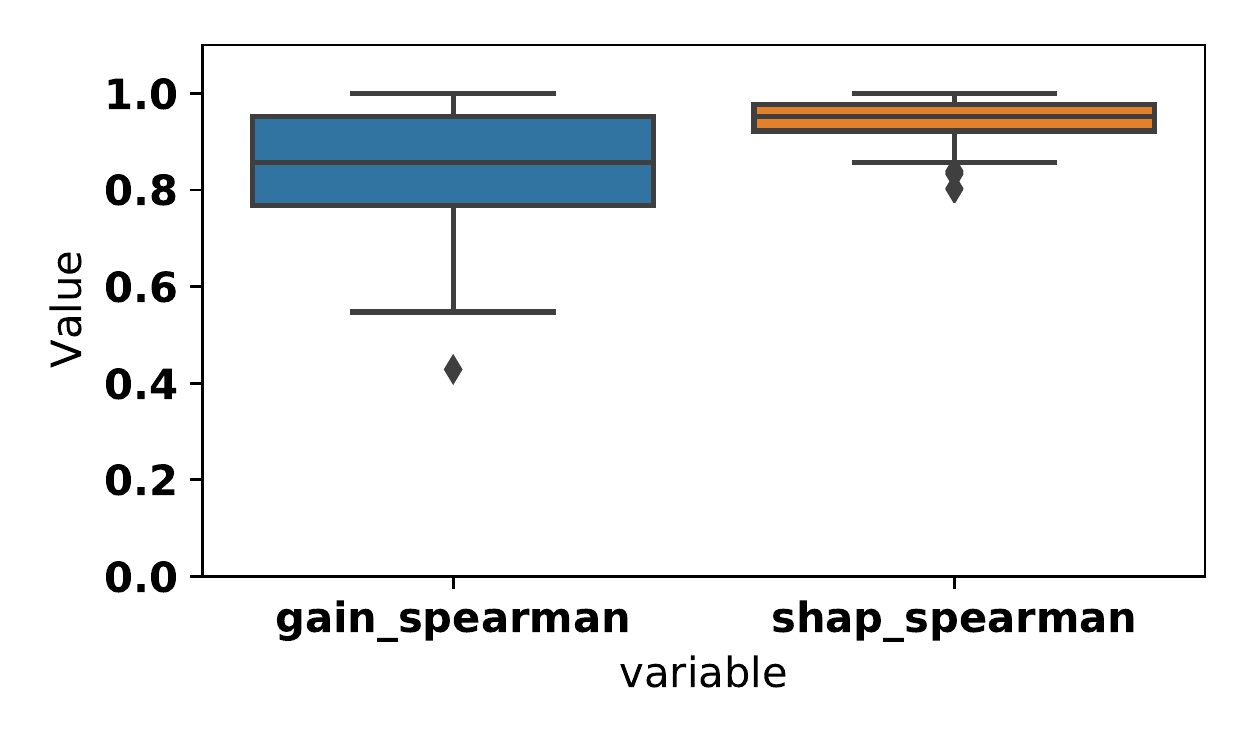} 
    \includegraphics[width=0.31\linewidth]{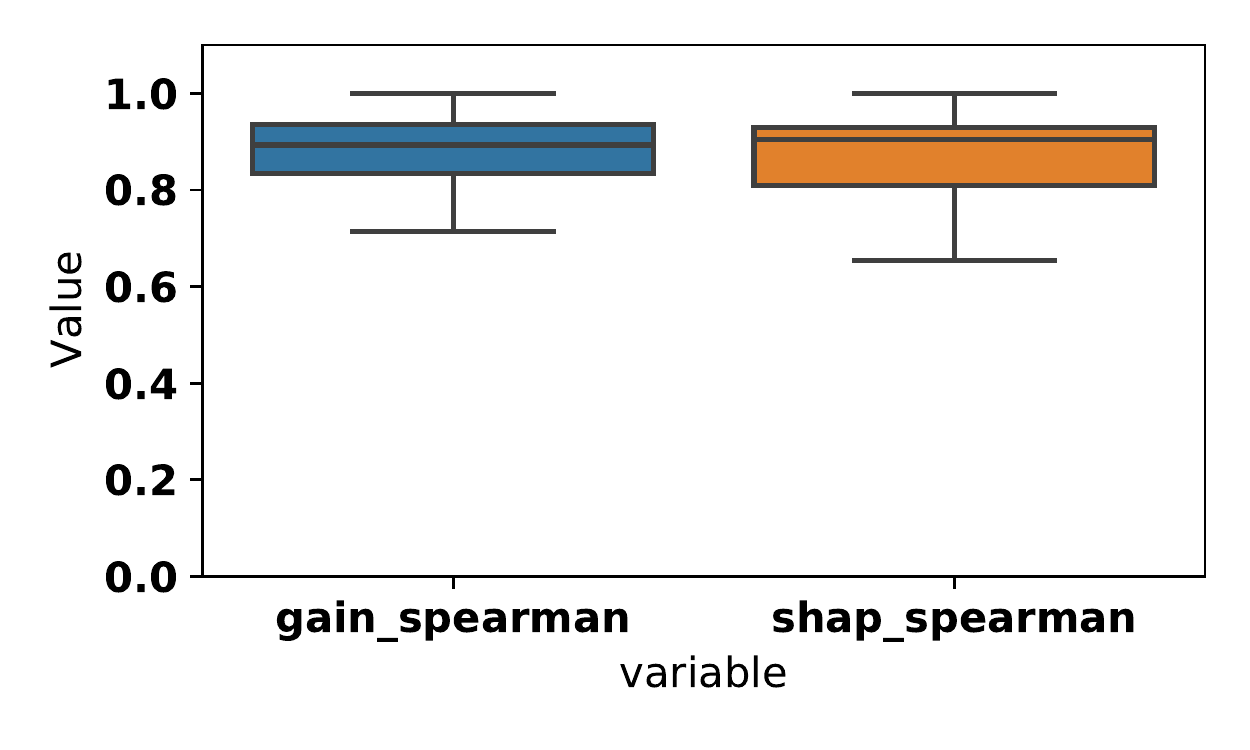} 
    \includegraphics[width=0.31\linewidth]{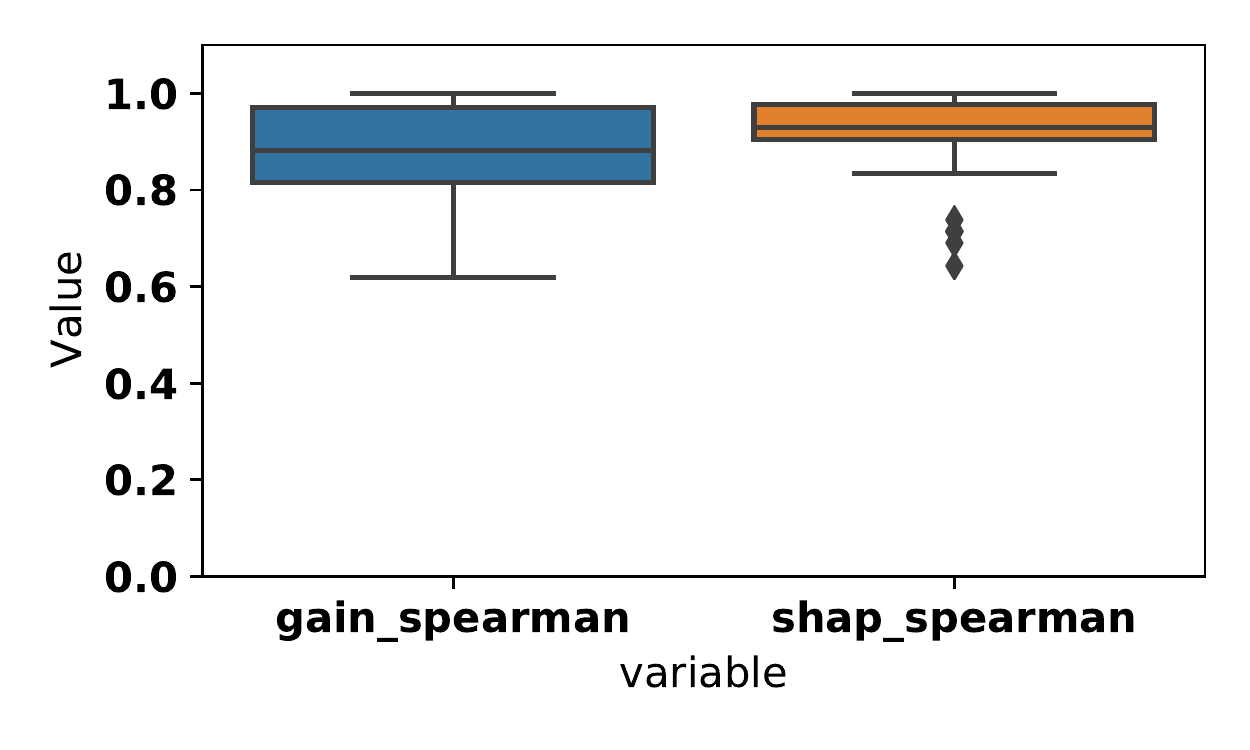} 
    
    Dataset: Auto MPG (\# Features: 7) \\
     XGBoost  \hspace{1.7cm} Gradient Boosting Machine \hspace{1cm} Random Forest \\ 
    \includegraphics[width=0.31\linewidth]{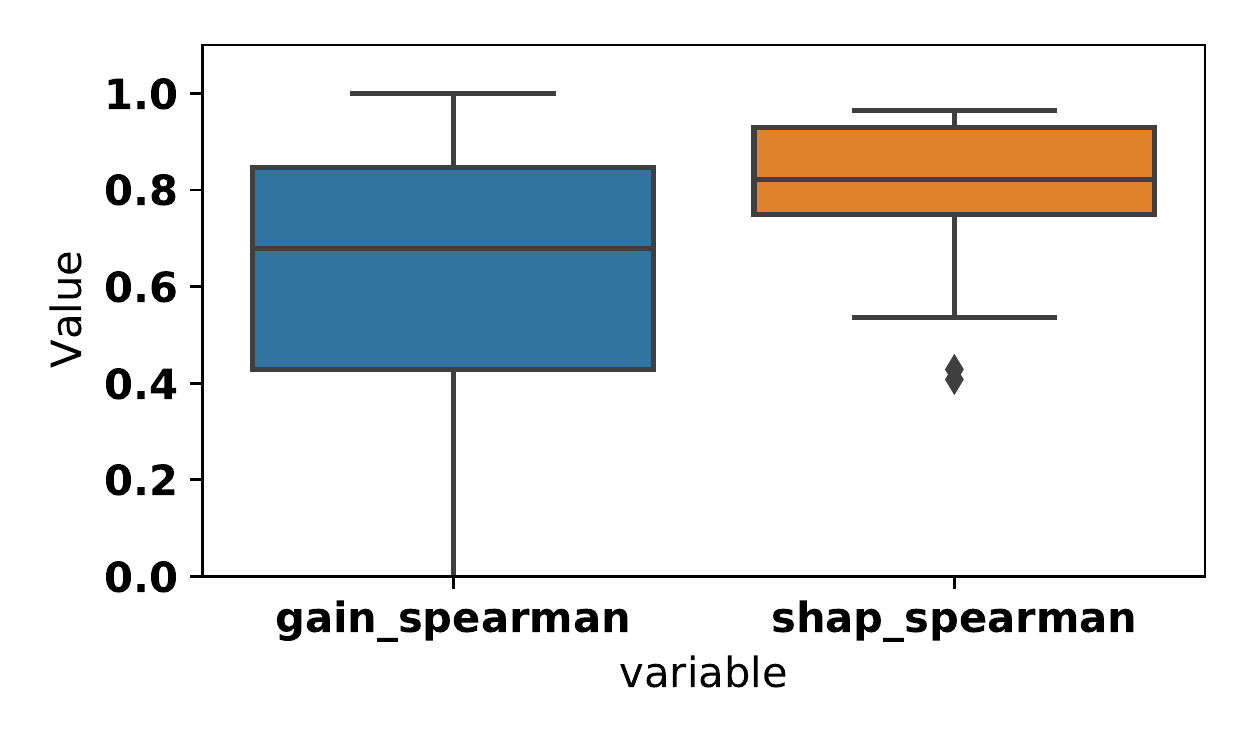}
    \includegraphics[width=0.31\linewidth]{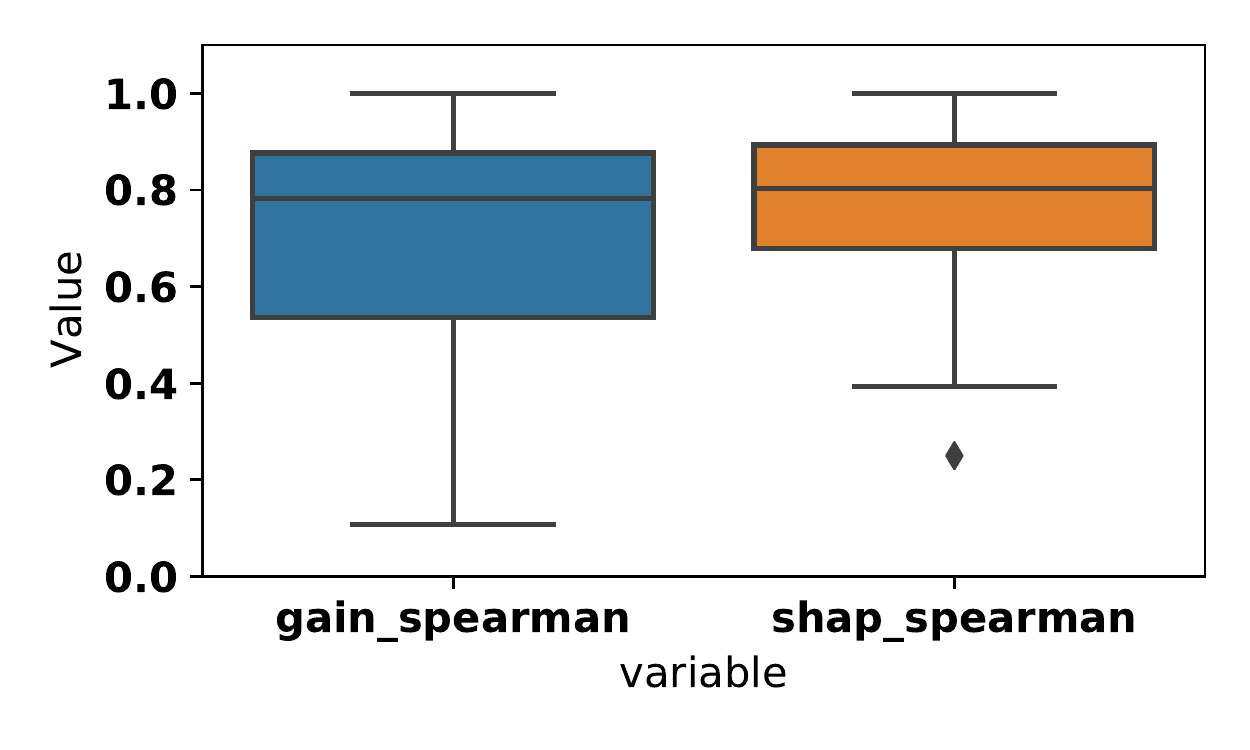}
    \includegraphics[width=0.31\linewidth]{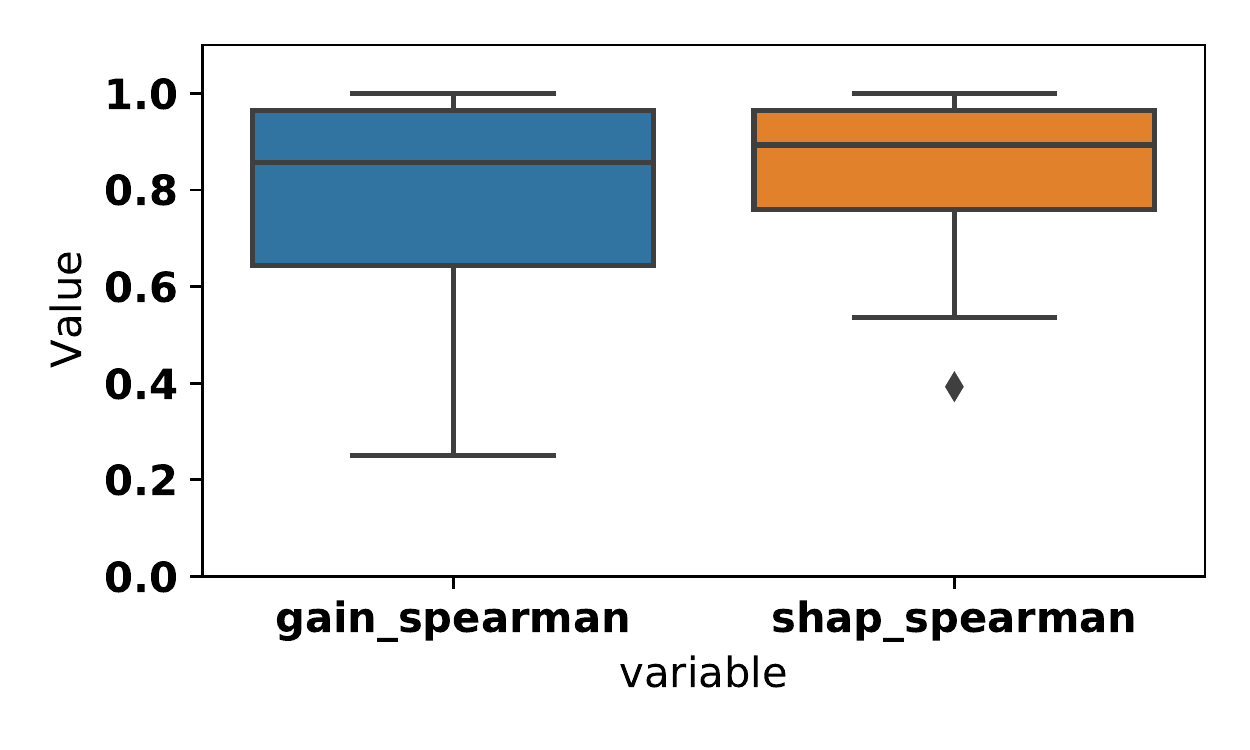}
    
    Dataset: Company Finance (\# Features: 892)\\
     XGBoost  \hspace{1.7cm} Gradient Boosting Machine \hspace{1cm} Random Forest \\ 
    \includegraphics[width=0.31\linewidth]{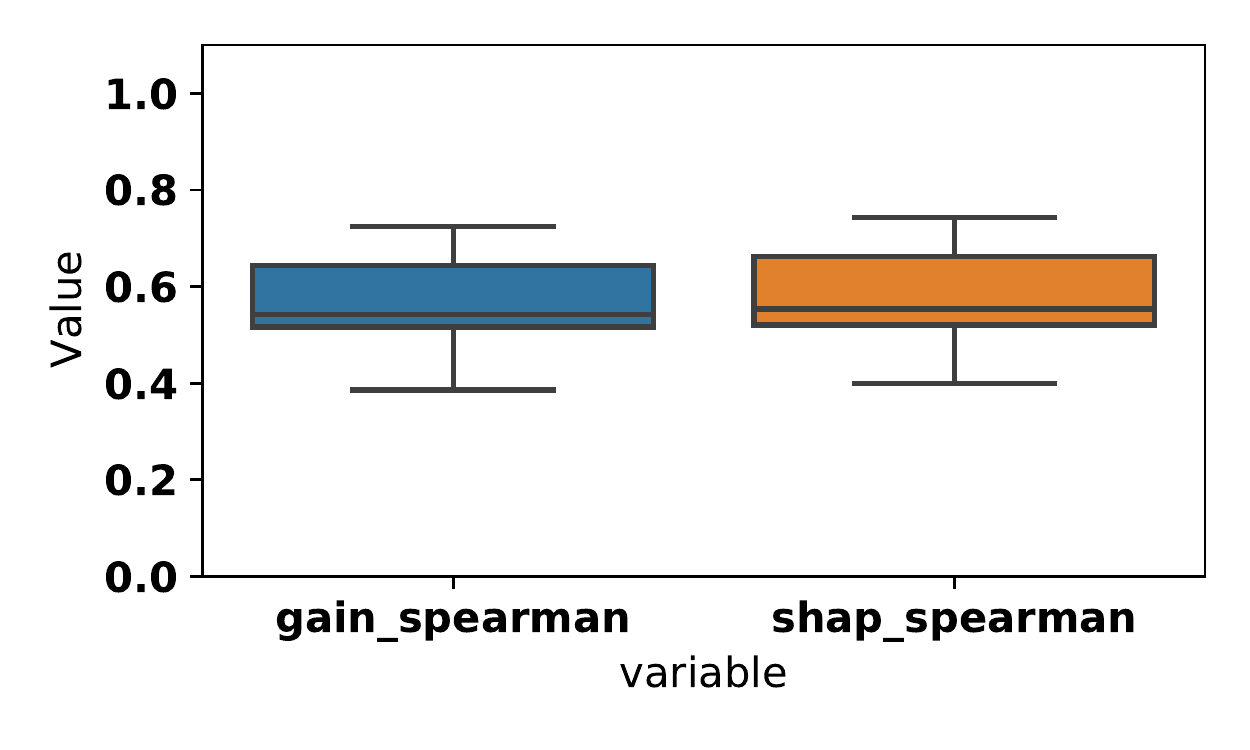}
    \includegraphics[width=0.31\linewidth]{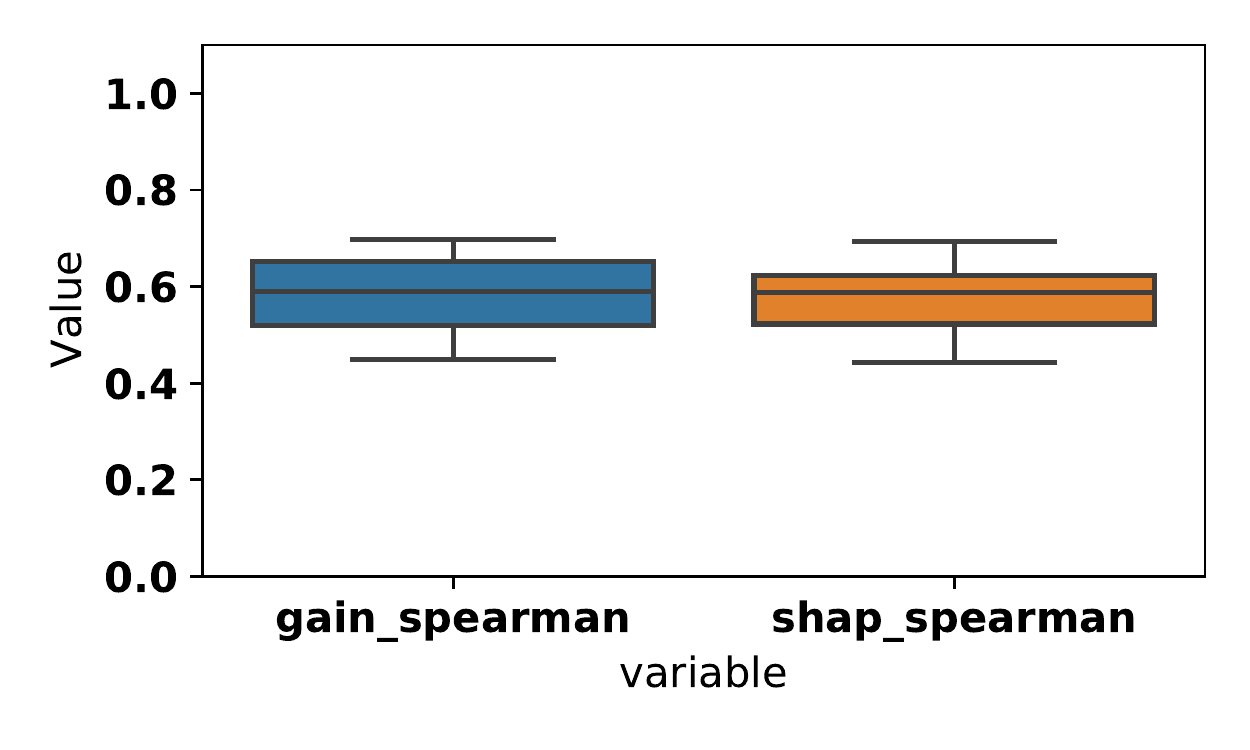}
    \includegraphics[width=0.31\linewidth]{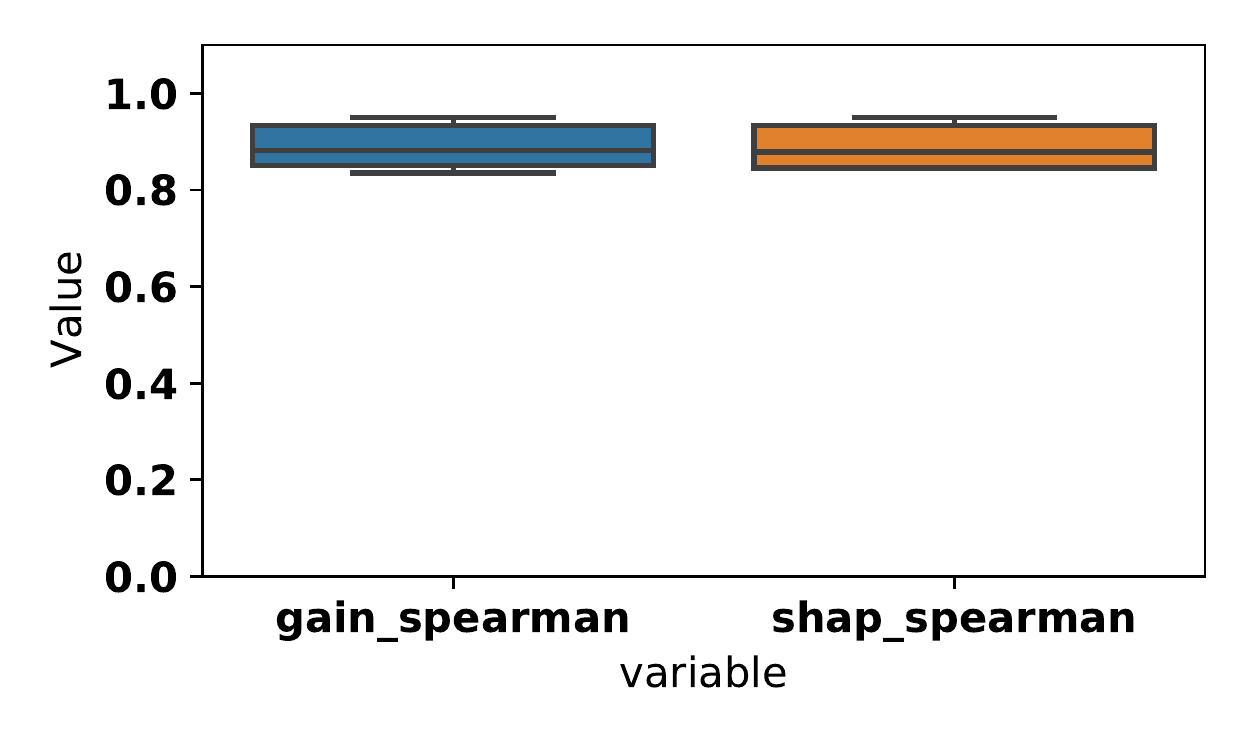}
    
    \caption{Correlation of feature importances (Blue: gain, Orange: SHAP) for XGBoost models trained on four real-world datasets with perturbations to the model's hyperparameter settings. SHAP is slightly more stable than gain for XGBoost.}
    \label{fig:real_world_diff_seed}
\end{figure}

In real world settings, we also notice a decrease in stability for gain and SHAP when models' hyperparameter settings are perturbed (Figure \ref{fig:real_world_diff_seed}). This is especially bold for Forest Fire dataset. On average, gain feature importances have around 60\% Spearman correlation whereas SHAP have around 90\% Spearman correlations in this dataset. 
SHAP tends to be more stable across the different real-world datasets, especially for XGBoost model as shown in Figure \ref{fig:real_world_diff_seed}, although this uplift is not as apparent in Gradient Boosting Machine and random Forest models.



\vspace{-.3cm}
\subsection{Summary of Results}
\label{results:summary}
We observe that there is a lack of accuracy with gain and SHAP feature importances even when there is no perturbation involved. In synthetic data with 5 features, the top feature is only ranked correctly around 40\% of the time. In addition to lack of accuracy, we also evaluate the lack of stability of these feature importances in various settings. 
We find that when inputs are perturbed, the correlations drop very low, both in synthetic and real-world datasets. When we perturb the models, especially by using different hyperparameter settings, correlation of feature importances can drop to 70-80\%. We find SHAP to be slightly more stable than gain in many cases, but both of their Spearman correlations still reduces to 60\% when low noise is added to the input.


\vspace{-.2cm}
\section{Discussion} \label{sec:discussion}
\vspace{-.2cm}
We set out to investigate the accuracy and stability of global feature importances for tree-based ensemble methods, such as random forest, gradient boosting machine, and XGBoost. We mainly look at two feature importance methods \emph{gain}, and SHAP. For both of these methods, we evaluate the accuracy in a simulated environment where true coefficients are known with and without noisy inputs. We also evaluate the stability of these methods in two directions, that is
 \begin{enumerate*}
    \item when inputs are perturbed, and
    \item when model settings are perturbed, either by initializing with a different random seed or by optimizing their hyperparameters with a different random seed.
\end{enumerate*}

\paragraph{Accuracy Analysis.} We find that SHAP tends to be better at accurately identifying top features compared to gain, although the overall accuracy of both is quite low especially when considering the ordering of all the features. 

\paragraph{Stability Analysis.} 

In our experiments, we find that SHAP is either equally or more stable when compared with gain. This is especially interesting as both gain and (Tree) SHAP feature importances investigated here use the innate structure of the trees. The difference lies on the fact that gain measures the feature's contribution to accuracy improvements or decreasing of uncertainty/variance whereas SHAP measures the feature's contribution to the predicted output.

\paragraph{Future Work.} There has been recent work on extending Shapley values to other cooperative game theory algorithms, such as the core \cite{yan2020if}. We will investigate this approach when a public implementation of this algorithm becomes available. In this study, we mostly focus on the stability of global features importance across the same model trained with perturbed hyperparameters/random seeds or inputs. Dong and Rudin recently suggest the idea of using a variable cloud importance, capturing the many good (but not necessarily the same) explanations coming from a group of models with almost equal performance \cite{dong2019variable}. In our future work, we will investigate the stability and usability of this methodology. We will also extend our analysis to new scenarios and datasets.  


\paragraph{Conclusion.} We investigate the accuracy and stability of global feature importances for tree ensemble methods. We find that even though SHAP in many cases can be more stable than gain feature importance, both methods still have limitations in terms of accuracy and stability and more work needs to be done to make them trustworthy. We hope that our paper will continue propel the discussion for trustworthy global feature importances and for the community to investigate this more thoroughly.

\bibliographystyle{splncs04}
\bibliography{reference}

\newpage
 \appendix

  \section{Determinism of XGBoost Feature Importances} \label{appendix:xgboost_deterministic}
 
  \begin{figure*}[h]
  \centering
  \includegraphics[width=0.49\linewidth]{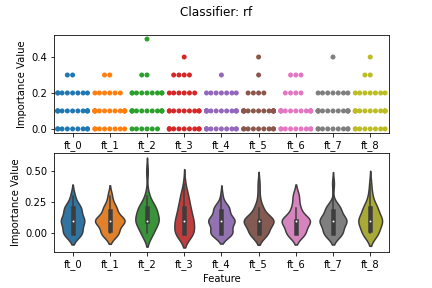}
  \includegraphics[width=0.49\linewidth]{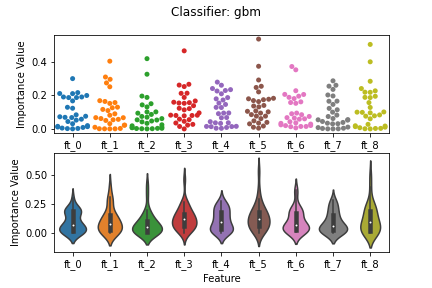} \\
  \includegraphics[width=0.49\linewidth]{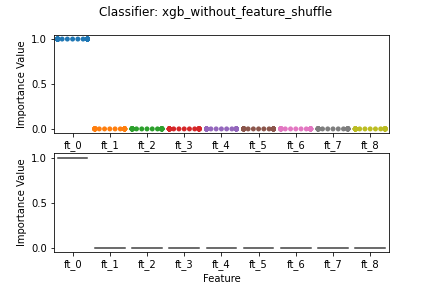}
  \includegraphics[width=0.49\linewidth]{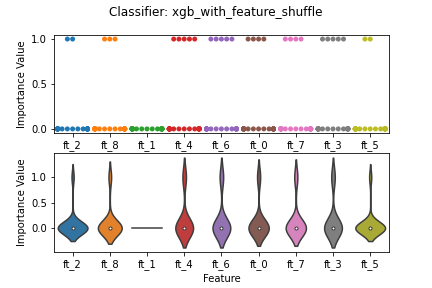}
  
  \caption{These plots show the distribution of feature importances across 10 redundant features for random forest (top left), gradient boosting (top right), XGBoost (bottom left), and XGBoost with feature shuffling (bottom right). XGBoost by its implementation is more deterministic compared to other methods at assigning feature importance. For the same hyperparameter with different seeds, when the features are redundant, it will always pick the first feature in order. With feature shuffling though, we are able to break this pattern a little bit. }
  \label{fig:feature_imp_stability}
\end{figure*}

\begin{figure*}[h]
  \centering
  \scriptsize
  Perturbation: Input (Low noise) \hspace{2.2cm} Model (random seeds)\\
  \includegraphics[width=0.4\linewidth]{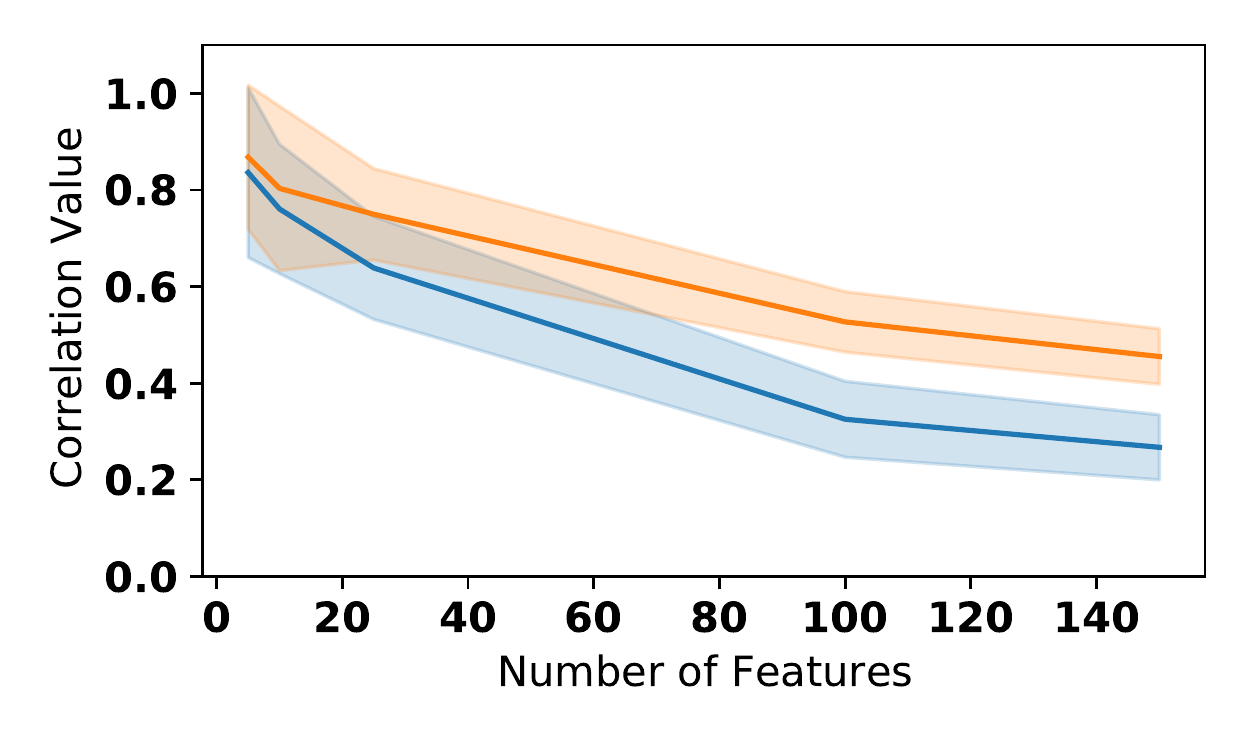}
  \includegraphics[width=0.1\linewidth]{images/simulation/legend.png}
  \includegraphics[width=0.4\linewidth]{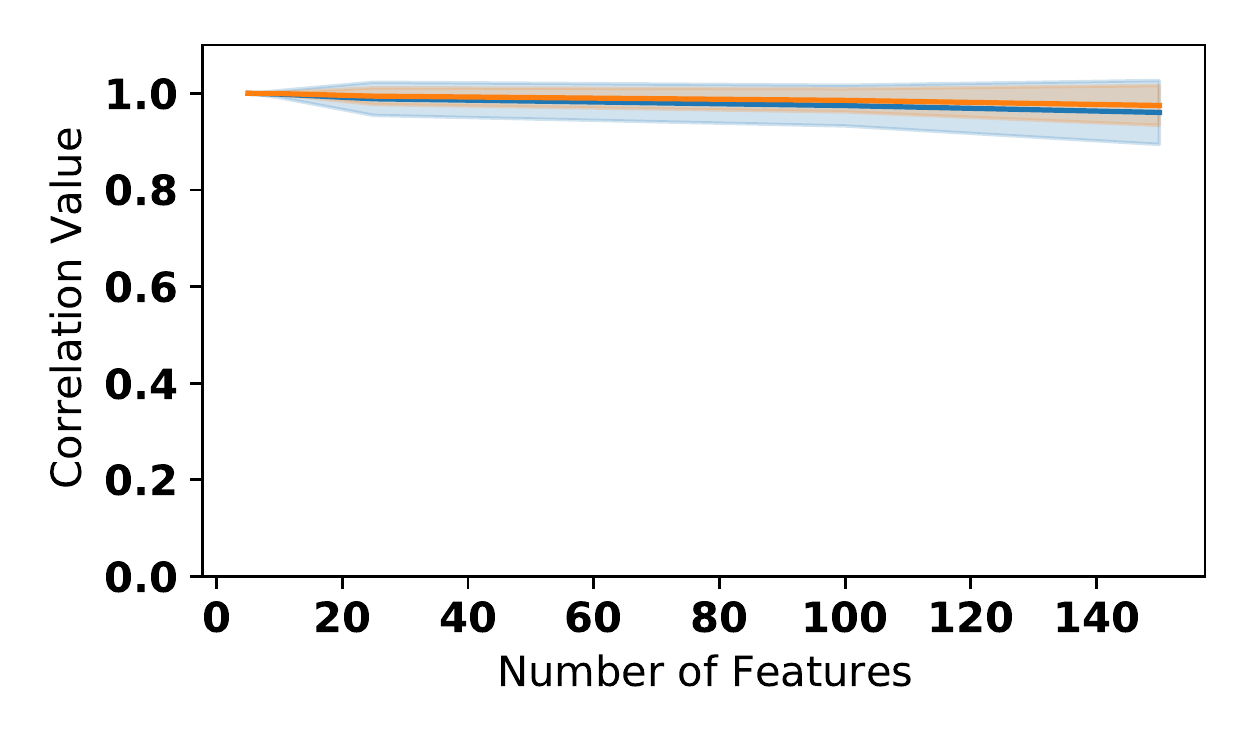} \\
  
  Perturbation: Model (hyperparameters)\hspace{1.7cm} Model (hyperparameters) \& input (low noise) \\
  \includegraphics[width=0.4\linewidth]{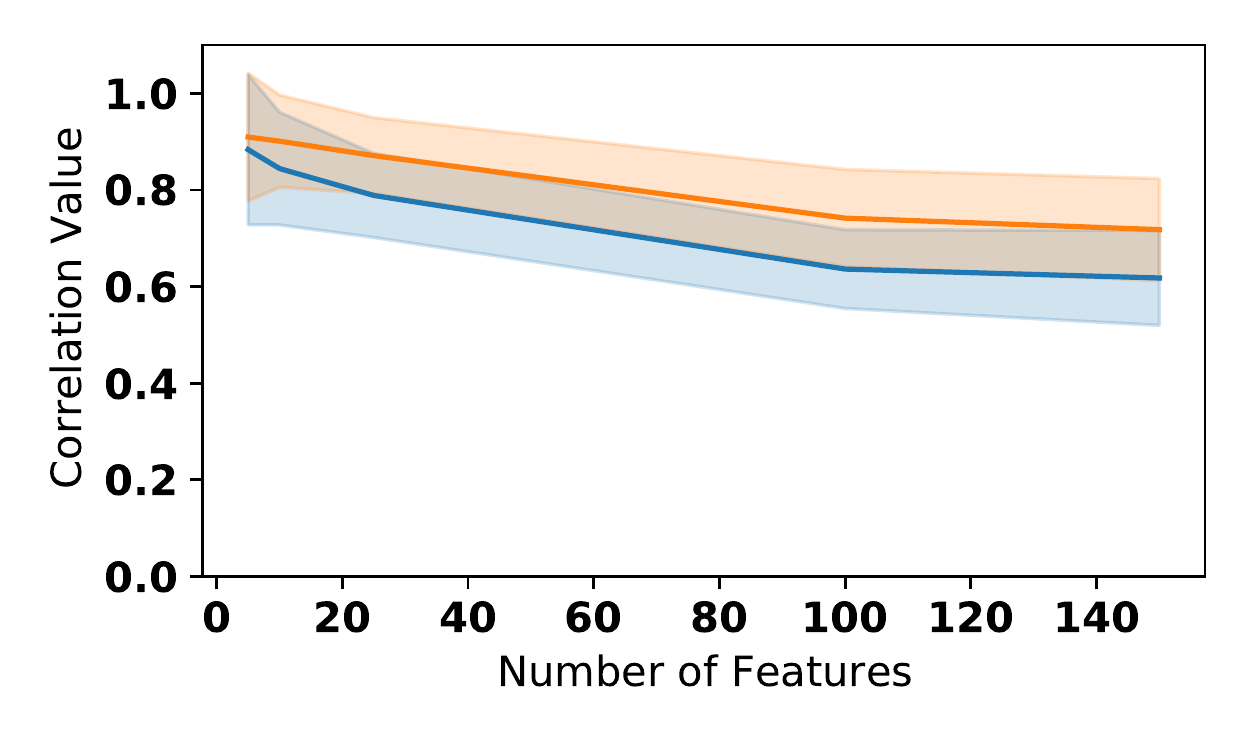}
  \hspace*{10ex}
  \includegraphics[width=0.4\linewidth]{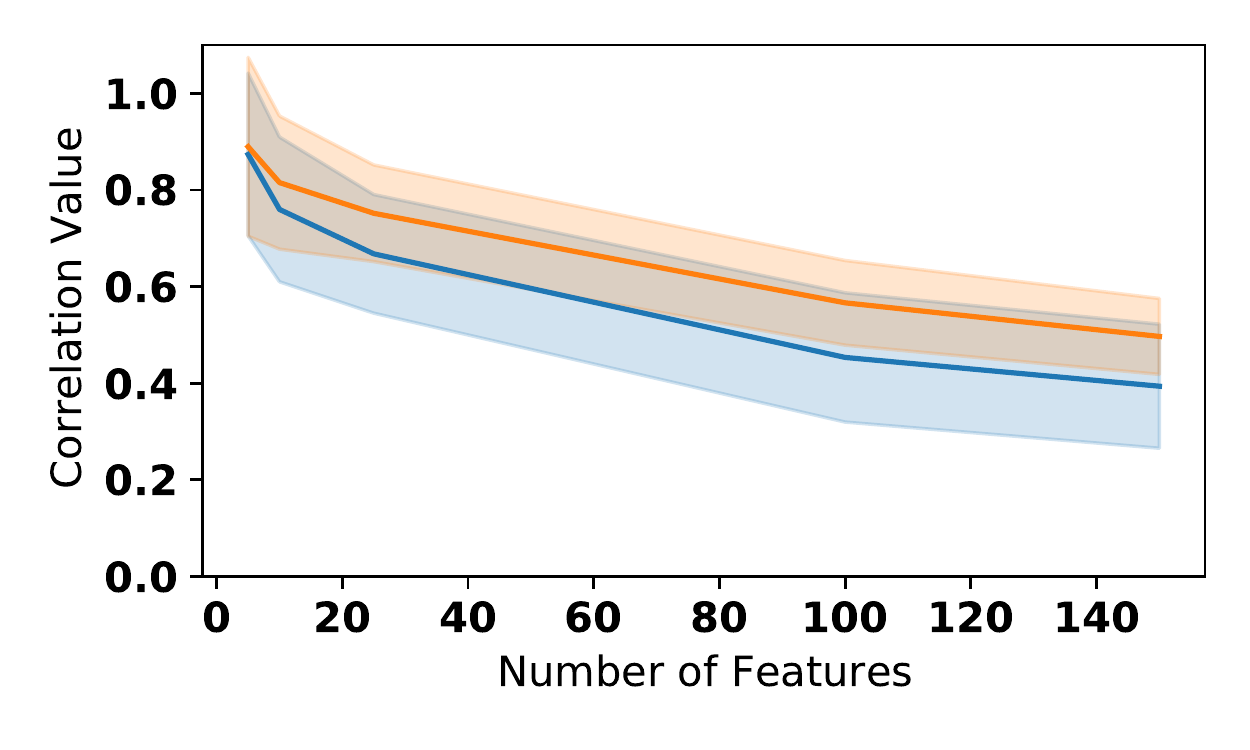}
  \caption{SHAP is more stable overall for XGBoost with shuffled features as can be seen on the plots above across input perturbation (low noise) experiments, model perturbations and both. Each row represents a different set of experiments with Spearman correlations of the default feature importance (Blue) and SHAP feature importance (Orange).}
  \label{fig:xgb_shuffled}

\end{figure*}

In this experiment, we simulate 1000 samples with 10 redundant features where each feature is equally important in predicting the target. Figure \ref{fig:feature_imp_stability} shows the distribution of the default feature importance in random forests, gradient boosting, and XGBoost across 30 iterations with different random seeds. As shown on the bottom left, XGBoost always assigns all importance to the first feature it saw no matter the random seed. When we shuffle the order of the features, we are able to break down this pattern (shown on bottom right). This is why on Figure \ref{fig:perturb_model}, there is a perfect correlation of importance for XGBoost initialized with different random seeds. With shuffled features, we still find SHAP to be more stable for XGBoost overall, although the correlation still decreases with higher number of features (See Figure \ref{fig:xgb_shuffled}).
 
\end{document}